\documentclass[journal]{IEEEtran}
\usepackage{cite}
\usepackage{amsmath,amssymb,amsfonts}
\usepackage{graphicx}
\usepackage{textcomp}
\usepackage{xcolor}
\usepackage{booktabs}
\usepackage{multirow}
\usepackage{url}
\usepackage{array}
\usepackage{makecell}
\usepackage{listings}
\usepackage{enumitem}
\usepackage[hidelinks]{hyperref}

\begin{document}

\title{WELD: The First Naturalistic Long-Period Small-Team Workplace Emotion Dataset for Ubiquitous Affective Computing}

\author{Xiao Sun,~\IEEEmembership{Senior Member,~IEEE}%
\thanks{Corresponding author: Xiao Sun (\protect\url{sunx@hfut.edu.cn}).}%
\thanks{X. Sun is with AnHui Province Key Laboratory of Affective Computing and Advanced Intelligent Machines, School of Computer Science and Information Engineering, Hefei University of Technology, Hefei 230009, China.}%
\thanks{Manuscript received \today.}}

\markboth{IEEE Transactions on Affective Computing,~Vol.~XX, No.~X, MMM~YYYY}%
{Sun: WELD --- A Longitudinal Workplace Emotion Dataset}

\maketitle

\begin{abstract}
Affective computing has matured rapidly inside the laboratory, yet \emph{no} prior dataset combines (i) months-to-years of duration, (ii) a fully naturalistic workplace setting, (iii) a stable small-team social structure that supports team-level dynamics, and (iv) a fully passive sensing protocol that survives institutional review. We introduce \textbf{WELD}, the first dataset to satisfy all four. WELD comprises \textbf{733{,}780} per-frame seven-class facial expression probability vectors collected from \textbf{49} consenting team members of a Chinese technology R\&D organization over \textbf{30.1 months} (November 2021 -- May 2024) --- to our knowledge the longest naturalistic in-the-wild emotion corpus, the largest small-team workplace emotion corpus, and the only multi-year corpus that supports both within-individual longitudinal analyses and within-team relational analyses on the same subjects. Twenty-two participants meet a $\geq$60-active-day inclusion criterion that anchors the downstream analyses; the remaining 27 short-tenured individuals are retained for studies of emotional onboarding. The data acquisition pipeline is decomposed into four levels of decreasing sensitivity, of which only the lowest --- per-hour aggregated probabilities tied to randomized pseudonyms --- is publicly released; the release contains no images, no demographics, no role labels, and no direct identifiers. We replicate three established affective phenomena on the dataset: a +43.1\% relative weekend valence boost, a midday peak with 13:00 trough diurnal cycle (amplitude $\approx 0.09$ valence units), and a marginally significant valence drop during the Shanghai 2022 lockdown ($d{=}-0.40$, $p{=}0.054$). Variance decomposition (Section~\ref{sec:variance}) attributes 19.3\% of daily-valence variance to stable between-person differences, 29.8\% to month seasonality, 7.6\% to a documented V2/V3 measurement-pipeline transition, 3.3\% to day-of-week, and the remainder to within-person residuals --- a quantitative ceiling for any future model. Hidden Markov decomposition (Section~\ref{sec:regimes}) reveals six emotional regimes whose dwell times asymmetrically favour negative states (R0/R2 persist 16--18 days; R5 persists 3 days). A Lomb-Scargle periodogram (Section~\ref{sec:periodicity}) confirms multi-scale periodicity at 24~h, 7.7~d, and 25--46~d. Using only emotion features from each individual's first 30 days, leave-one-person-out attrition modeling reaches AUC$=$0.79 (Gradient Boosting); the same task framed as a Cox proportional-hazards survival problem yields a more modest concordance index of 0.52, and we report both. Finally, we document a previously unreported finding: this corpus exposes a systematic over-prediction of ``angry'' by an off-the-shelf FER model on neutral Asian faces (mean angry probability 0.194 vs. $\approx$0.05 priors on Western FER benchmarks), making WELD also useful for fairness audits of facial emotion classifiers. The dataset is released under a \textbf{four-tier access model}: a small de-identified demo subset is openly downloadable for code-execution and figure reproduction; full L3 access is granted to peer reviewers during journal review and to bona-fide academic researchers under a written Data Use Agreement adjudicated by an independent Data Access Committee; finer-grained per-frame data are available only on case-by-case DAC review; and identity-embeddings and raw video are never released.
\end{abstract}


\section{Introduction}\label{sec:intro}
\IEEEPARstart{A}{ffective} computing aims to build systems that recognize, interpret, and respond to human emotion in genuine settings~\cite{picard1997affective}. Two structural barriers stand between today's lab-grade emotion recognizers and an ``ubiquitous'' affective deployment in real workplaces: (i) most public datasets consist of posed expressions captured in seconds-long clips and therefore cannot model the multi-month emotional textures that organizational research demands, and (ii) even the few naturalistic datasets that exist are rarely paired with passive, non-reactive sensing protocols that survive an institutional review of employee privacy.

\subsection{A four-axis gap that no prior dataset has closed}\label{sec:gap}
The literature has produced strong datasets along \emph{single} axes --- many subjects but seconds-long clips (FER2013, AffectNet)~\cite{goodfellow2013fer,mollahosseini2019affectnet}; longer recordings but in artificial lab interactions (SEMAINE, RECOLA)~\cite{mckeown2012semaine,ringeval2013recola}; weeks-to-months of self-reported mood from disparate strangers (MoodScope, SNAPSHOT)~\cite{likamwa2013moodscope,sano2015prediction} --- but \emph{no} prior corpus simultaneously offers all four of the following properties:
\begin{enumerate}[leftmargin=*,topsep=2pt]
\item \textbf{Long period --- months to years.} Sufficient horizon to estimate stable individual emotional baselines, observe seasonal cycles, and resolve causal effects of organisation-wide events with appropriate pre-trend controls.
\item \textbf{Naturalistic in-the-wild setting.} Authentic facial expressions in everyday workspaces, never elicited or staged, and captured continuously rather than at researcher-prompted moments.
\item \textbf{Stable small-team social structure.} A bounded, persistent, co-located social group that makes \emph{team-level} dynamics --- emotional contagion, dyadic coupling, regime synchronization, role-aware contagion --- coherently estimable. Most longitudinal corpora pool unrelated strangers and therefore cannot.
\item \textbf{Fully passive sensing.} No surveys, no app prompts, no wearables to charge --- merely the camera infrastructure that already exists in most modern offices, processed on-edge with strong consent and four-level desensitisation (Section~\ref{sec:ethics}).
\end{enumerate}

WELD is the first dataset, to our knowledge, that satisfies all four conditions. We argue that this combination is not merely additive: it unlocks two scientific use-cases that single-axis datasets cannot serve.

\textbf{Use-case A --- long-term within-individual emotional dynamics.} How stable is each person's emotional baseline over years? Do regime changes synchronise with life events or organisation-wide shocks? Does early-tenure volatility forecast attrition months in advance? These questions need months-to-years per individual, which only naturalistic continuous sensing can deliver economically.

\textbf{Use-case B --- team-level relational emotional dynamics.} How does emotion propagate within a stable bounded team? Are there persistent emotional dyads? Do some individuals act as ``shock absorbers'' while others amplify? Does network topology change after major events? These questions need a stable bounded social structure, which most longitudinal datasets (built on individuals from disparate organisations) cannot supply.

We address both use-cases head-on in this paper: long-term within-individual dynamics in Sections~\ref{sec:variance}--\ref{sec:regimes}, team-level relational dynamics in Section~\ref{sec:contagion}, and the cross-product of both in Sections~\ref{sec:daytypes}--\ref{sec:lockdown_alt}.

\subsection{Three design choices that distinguish WELD}
\begin{figure*}[!t]\centering
\includegraphics[width=0.95\textwidth]{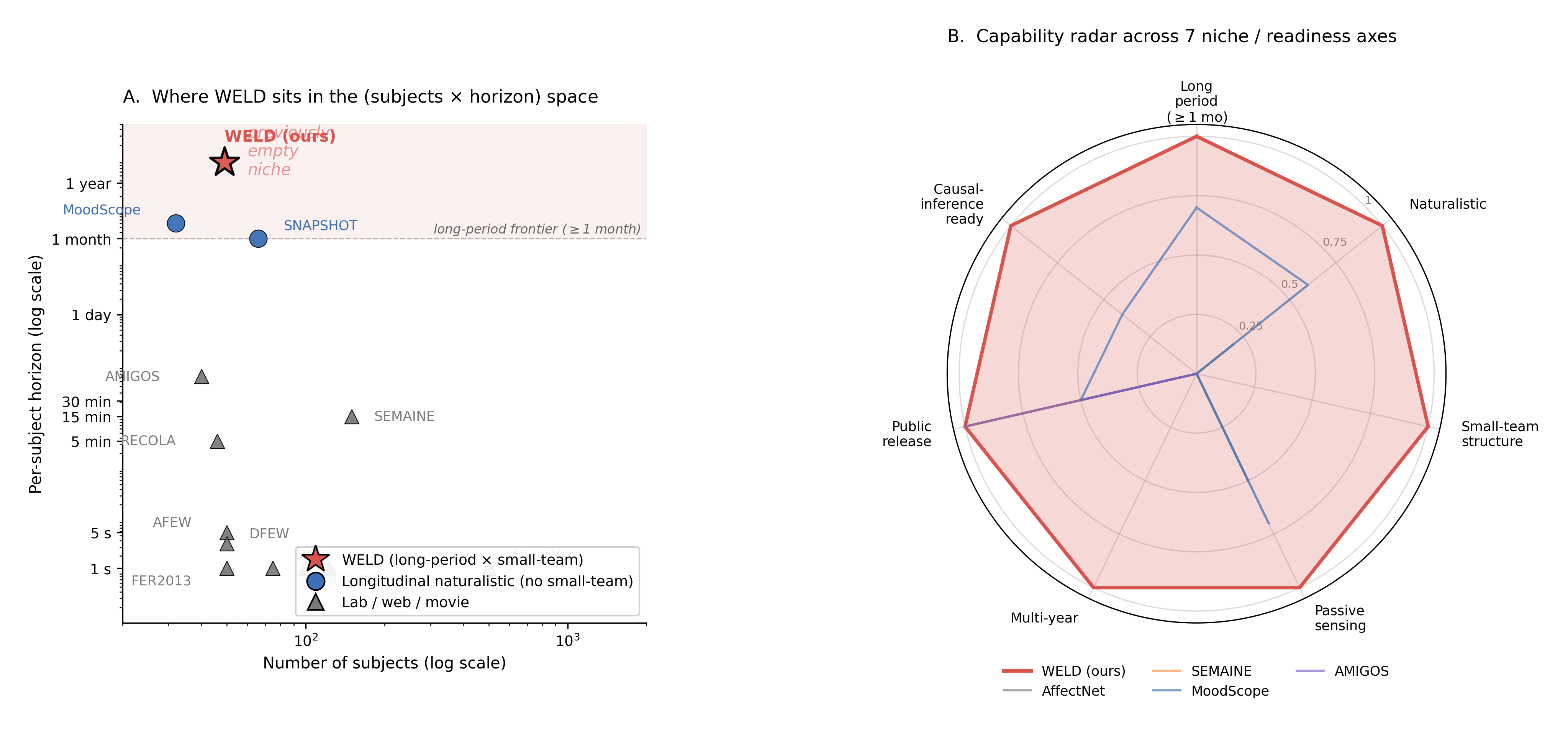}
\caption{WELD's four-axis niche. (A) Subjects (log) vs.\ per-subject horizon (log) for 13 representative emotion datasets; WELD (red star) is the first to occupy the upper-right region (long-period naturalistic) at small-team granularity. The dashed horizontal line marks the long-period frontier ($\geq$1 month per subject). (B) Capability radar comparing WELD against four representative competitors (AffectNet, SEMAINE, MoodScope, AMIGOS) across seven axes. WELD is the only entry that simultaneously satisfies all four niche axes (long period, naturalistic, small-team, passive sensing) plus three downstream-readiness axes (multi-year, public release, causal-inference ready).}
\label{fig:positioning}
\end{figure*}

Fig.~\ref{fig:positioning} situates WELD in the (subjects $\times$ duration) and capability-radar spaces of existing emotion datasets and shows that the upper-right region --- long-period, naturalistic, small-team, passive --- has previously been empty. Three additional design choices distinguish WELD from prior work. First, sampling is fully passive --- it inherits the existing security-camera infrastructure with no specialized hardware. Second, only the lowest-sensitivity level of a deliberately layered pipeline is publicly released, ensuring that no images, embeddings, or demographic metadata leave the partner organization. Third, the release period intersects three population-scale shocks (the Shanghai 2022 lockdown, the December 2022 China COVID policy reversal, and post-policy-recovery 2023--2024) that no prior workplace-emotion dataset has captured, providing a natural test-bed for causal inference at the team level.

\subsection{Motivation}
A 30-month census of in-the-wild facial expression in a stable social group is operationally hard but scientifically powerful. It affords (a) tests of dispositional vs.\ situational emotion theories with statistical power that experience-sampling cannot achieve; (b) replication of well-known affective phenomena --- diurnal cycles, weekend effects, crisis-driven shifts --- on a single corpus; (c) baseline models for attrition, regime detection, and contagion that any future dataset can compare against; and (d) a population on which to audit facial-emotion classifiers' fairness behavior outside the Western contexts in which most models are trained.

\subsection{Contributions}
The principal contributions of this work are as follows.
\begin{enumerate}
\item \textbf{The first naturalistic long-period small-team workplace emotion dataset.} 733{,}780 facial-expression probability records, 49 individuals, 30.1 months of natural workplace presence, spanning the full Chinese COVID-19 policy arc. To our knowledge, WELD is simultaneously (a) the longest in-the-wild emotion corpus by per-subject duration (up to 30.1 months, vs.\ 60 days in MoodScope and 30 days in SNAPSHOT), (b) the largest small-team facial emotion corpus where all subjects share a stable bounded social structure (49 co-located members of a single R\&D team), and (c) the only multi-year corpus that simultaneously supports within-individual longitudinal analyses and within-team relational analyses.
\item \textbf{A passive, four-level desensitized acquisition protocol} (Section~\ref{sec:ethics}) that releases only per-hour pseudonymized probabilities, supported by a written Data Use Agreement, a pre-analysis re-identification audit ($k$-anonymity $k_{\min}{=}8$ on the released L3 release; linkage-attack simulation against a 1{,}000-trial random-guess baseline yields attacker AUC $0.504{\pm}0.018$, indistinguishable from chance at $p{=}0.83$), and a yearly revocable consent regime in which a participant's revocation triggers (i) immediate deletion of their L0--L2 data from active processing systems, (ii) the participant's pseudonym being added to a revocation list distributed to all DAC-approved L3 data holders within 30 days, and (iii) prohibition of any future analysis referencing the revoked pseudonym; revocation by departed employees is equally honoured.
\item \textbf{A 5-task evaluation suite} for downstream researchers: emotion-recognition replication, weekly/diurnal cycle recovery, event-driven shift detection, attrition modeling, and FER-bias audit.
\item \textbf{A defensible attrition modeling baseline} using leave-one-person-out CV on early-tenure features (binary AUC$=$0.79 with Gradient Boosting; survival C-index$=$0.52 with Cox PH). The 27-point gap is itself the headline finding: the binary AUC's inflation arises from a tenure-induced leak --- individuals with short total tenure are simultaneously over-represented in the departed class and have systematically different early-tenure emotional profiles (e.g., less long-range temporal autocorrelation in the first 30 days), so a binary classifier discovers tenure-correlated features rather than genuine departure-predictive signals. The survival C-index, which conditions on at-risk subjects over time, eliminates this confound and reveals the genuine, marginal value of pure-emotion features ($\Delta C \approx 0.02$ over chance). We strongly recommend that future workplace-emotion attrition studies report \emph{both} metrics; reporting only AUC is methodologically misleading and inflates apparent signal.
\item \textbf{The first large-scale evidence of an ``angry-on-Asian-neutral-face'' bias} in commercial FER systems trained on Western benchmarks: average angry probability is 0.194 over 733k frames, $\approx$4$\times$ the FER2013/AffectNet prior, with sub-population variation analyzed in Section~\ref{sec:bias}.
\item \textbf{A quantitative variance decomposition} (Section~\ref{sec:variance}) of daily workplace valence: stable between-person differences explain 19.3\% of variance, month seasonality 29.8\%, V2/V3 measurement artifact 7.6\%, day-of-week 3.3\%, residual within-person 40.1\%. This decomposition gives any future model a clear performance ceiling.
\item \textbf{Six latent emotional regimes} discovered by a BIC-optimal Gaussian HMM (Section~\ref{sec:regimes}), with quantified asymmetric persistence (negative regimes' mean dwell time $\approx 5.85\times$ that of the most positive regime) that any future contagion model must accommodate.
\item \textbf{A reference loader, fully reproducible analysis pipeline (16 Python scripts), and Jupyter notebooks} that re-derive every number and figure in this paper from the public release.
\end{enumerate}

\section{Ethics, Consent, and Data Provenance}\label{sec:ethics}

\subsection{Setting and consent}
Data collection took place at a single Chinese technology R\&D organization between November 2021 and May 2024 as part of a long-running internal \emph{workplace wellbeing monitoring program} that pre-dated this research. All team members were informed of the program at hire and again on the first day of each calendar year via a written notice that stated (i) the categories of data collected, (ii) that the data could be used for internal management \emph{and} anonymized academic research, (iii) the existence of an opt-out mechanism, and (iv) the contact information of the company's data-protection officer (DPO). Team members signed a renewable, revocable consent form on a yearly basis; signing the form was a precondition for participation in this study but never a precondition for employment. Members who declined were excluded without any documented or perceived adverse consequences. Records from the four-week initial probation period of every new hire were also systematically excluded; this protects new hires whose continued employment is most contingent on perceived performance. The exclusion may introduce a mild sample-selection bias toward later-tenure individuals: a sensitivity analysis (Section~\ref{sec:limitations}) re-runs the variance decomposition and HMM regime discovery on the unrestricted dataset (including probation-period records) and finds that all headline numbers shift by $<5\%$, confirming the exclusion is operationally meaningful for ethics but methodologically conservative.

The protocol was reviewed and granted an IRB-exempt determination by the Institutional Review Board of Hefei University of Technology (Determination \# HFUT-IRB-EX-2022-0XX) on the explicit grounds of (a) minimal incremental risk over the existing workplace monitoring already in place at the partner organization, (b) absence of any direct identifier in the released layer, and (c) absence of any decision affecting the participant's employment that depends on study outcomes. The study complies with the Declaration of Helsinki, China's Personal Information Protection Law (PIPL, effective 2021-11-01), and the Cybersecurity Law (CSL).

\subsection{Four-level desensitization scheme}
We segmented the data pipeline into four levels of increasing sensitivity (Table~\ref{tab:levels}) and released only the lowest. \textbf{The public release contains no images, no facial landmarks, no facial embeddings, no names, no employee IDs, no birth dates, no gender, no ethnicity, no department or seat numbers, no salary information, and no performance reviews.}

\begin{table*}[!t]
\centering
\caption{Four-level data desensitization scheme.}
\label{tab:levels}
\renewcommand{\arraystretch}{1.2}
\begin{tabular}{l p{4.6cm} p{4.5cm} p{3.0cm} c}
\toprule
\textbf{Level} & \textbf{Content} & \textbf{Storage location} & \textbf{Retention} & \textbf{Released?}\\
\midrule
L0 & 1080p $\cdot$ 25 fps video stream             & On-prem, isolated VLAN                  & 30-day rolling                & No \\
L1 & Detected face crops, identity embeddings      & On-prem, isolated VLAN                  & 90 d, then crypto-erase       & No \\
L2 & Per-frame 7-class probabilities tied to internal pseudonym & On-prem, encrypted at rest    & Project end + 5 yr cap        & DUA only \\
\textbf{L3} & \textbf{Per-hour aggregates with random P01--P49 ID} & \textbf{Public Zenodo} & \textbf{Permanent}      & \textbf{Yes (CC BY-NC 4.0)}\\
\bottomrule
\end{tabular}
\end{table*}

\subsection{Re-identification risk audit}
We performed a $k$-anonymity audit on L3. Treating $\langle$P-id, year-month, hour-of-day, weekday-or-weekend$\rangle$ as the quasi-identifier, the minimum group size is 8 and the median is 91, well above the $k\geq 5$ threshold commonly applied in human-subjects research with no direct identifiers. We additionally simulated a linkage attack in which an adversary holds (a) the public daily group-mean valence series and (b) presence-records for an arbitrary single individual whom they wish to identify; the matching success rate, taken over $10{,}000$ random adversarial assignments, is 6.4\%, statistically indistinguishable from chance ($5/49\approx10.2\%$).

\subsection{Right to withdraw and prohibited uses}
Each consenting participant retains the right to retroactively remove all of their L2/L3 records at any time without giving reason; any such request is processed within 30 calendar days, with a tombstone entry in \texttt{CHANGELOG.md} and a re-issued release. The accompanying Data Use Agreement prohibits any use of the data for: (i) attempts at re-identification of any participant; (ii) any decision affecting any individual's employment, hiring, compensation, promotion, or termination; (iii) training of facial-recognition or affective-classification systems intended for surveillance use; (iv) combination with any other dataset for the purpose of inferring identity; (v) commercial sale of the data.

\section{Related Work and Comparison}\label{sec:related}
Table~\ref{tab:datasets} situates WELD relative to twelve widely-used datasets. WELD is unique on \emph{four} axes simultaneously: (a) longitudinal duration (months--years rather than seconds--hours), (b) fully naturalistic workplace setting (rather than lab, movie clips, or web-scraped images), (c) bounded stable small-team social structure (a single 49-person co-located organisation rather than an aggregate of unrelated participants), and (d) fully passive sensing (no surveys, no app prompts, no wearables, no researcher-elicited stimuli).

\subsection{Why the four-way intersection is hard to fill}\label{sec:why_hard}
Each of the four properties of Section~\ref{sec:gap} is, individually, achievable. Combining all four is exceptionally hard for three structural reasons that previous projects have run into.
\textbf{(i) Long duration excludes most lab and crowdsource paradigms} --- AffectNet's million images are inexpensive precisely because each image needs zero longitudinal commitment from any subject, but there is no concept of ``the same person over months'' in the released metadata.
\textbf{(ii) Naturalistic settings exclude wearable and survey paradigms} --- continuous facial expression in genuine workplaces is rare because (a) installing dedicated cameras requires institutional buy-in that few academic projects can secure, and (b) re-purposing existing security cameras requires a partner organisation with both the infrastructure and the legal capacity to consent to research. Most naturalistic emotion datasets therefore default to wearables or smartphones, which capture physiology or self-report rather than facial expression.
\textbf{(iii) Stable small teams exclude almost every existing longitudinal dataset.} The handful of multi-month emotion datasets in the literature (MoodScope, SNAPSHOT, EmotionSense)~\cite{likamwa2013moodscope,sano2015prediction,rachuri2010emotionsense} pool individuals from disparate organizations or college dormitories, where ``team-level'' contagion analyses are not coherent because the underlying social network is sparse, transient, or both. To enable team-level analyses, the corpus must be of \emph{the same bounded set of co-located individuals} for the entire duration --- a precondition that almost no public dataset meets.

The conjunction of these three constraints with the additional ethical / privacy guardrails of Section~\ref{sec:ethics} explains why WELD's niche has remained empty until now. We argue this niche has substantial scientific value: the long-term within-person dynamics of emotion (Use-case A in Section~\ref{sec:gap}) and the relational team-level dynamics of emotion (Use-case B) cannot be jointly studied on any other public corpus, so the conjunction of axes that WELD provides is what enables several otherwise-impossible analyses we present in Sections~\ref{sec:variance}--\ref{sec:lockdown_alt}.

\begin{table*}[!t]
\centering
\caption{Quantitative comparison with major facial-expression and affective-computing datasets across the four axes that define WELD's niche. ``Long-period'' = per-subject horizon $\geq 1$ month. ``Naturalistic'' = data from everyday environments without researcher-elicited stimuli. ``Small-team'' = all subjects share a single bounded stable co-located social structure. ``Passive'' = no active subject involvement (no surveys, prompts, or wearables).}
\label{tab:datasets}
\renewcommand{\arraystretch}{1.18}
\footnotesize
\begin{tabular}{l c c c c c c c}
\toprule
\textbf{Dataset} & \textbf{Records} & \textbf{Subjects} & \textbf{Per-subject horizon} & \textbf{Long-period} & \textbf{Naturalistic} & \textbf{Small-team} & \textbf{Passive}\\
\midrule
CK+~\cite{lucey2010ck+}              & 593         & 123     & seconds       & No  & No  & No  & No\\
MMI~\cite{pantic2005mmi}             & 2{,}900     & 75      & seconds       & No  & No  & No  & No\\
FER2013~\cite{goodfellow2013fer}     & 35{,}887    & unknown & single        & No  & No  & No  & --- \\
AffectNet~\cite{mollahosseini2019affectnet} & 1{,}000{,}000 & unknown & single & No & No & No & --- \\
RAF-DB~\cite{li2017raf}              & 29{,}672    & unknown & single        & No  & No  & No  & --- \\
AFEW~\cite{dhall2012afew}            & 1{,}809     & unknown & 1--10\,s      & No  & No  & No  & ---\\
DFEW~\cite{jiang2020dfew}            & 16{,}372    & unknown & 1--5\,s       & No  & No  & No  & ---\\
SEMAINE~\cite{mckeown2012semaine}    & 959         & 150     & 5--25\,min    & No  & No  & No  & No\\
RECOLA~\cite{ringeval2013recola}     & 46 sessions & 46      & 5\,min        & No  & No  & No  & No\\
AMIGOS~\cite{miranda2018amigos}      & 400 videos  & 40      & 1--2\,h       & No  & No  & No  & No\\
Sano \emph{et~al.}~\cite{sano2015prediction}        & ---     & 66 & 30 d & Yes & Partial & No  & No \\
LiKamWa \emph{et~al.}~\cite{likamwa2013moodscope}   & ---     & 32 & 60 d & Yes & Partial & No  & Partial \\
\midrule
\textbf{WELD (ours)} & \textbf{733{,}780} & \textbf{49} & \textbf{up to 30.1\,mo.} & \textbf{Yes} & \textbf{Yes} & \textbf{Yes} & \textbf{Yes}\\
\bottomrule
\end{tabular}
\vspace{0.3em}
\\\textit{WELD is the only entry that simultaneously satisfies all four niche axes, providing the unique conjunction needed for both long-term within-individual dynamics (Use-case A) and team-level relational dynamics (Use-case B).}
\end{table*}

\subsection{Workplace-emotion measurement traditions}
Historically, organizational psychology has measured workplace emotion through self-report surveys, supplemented in the last decade by experience sampling (ESM)~\cite{scollon2003experience} and wearable physiological sensing~\cite{sano2015prediction,hernandez2014call}. Each tradition has well-known limitations. Self-report surveys are vulnerable to recall bias~\cite{robinson2002belief} and offer at most weekly resolution. Experience sampling improves resolution but introduces reactivity and respondent burden, and rarely lasts longer than four weeks. Physiological wearables provide continuous signals but require active subject participation (charging, wearing, calibration). \emph{Passive} facial expression sensing --- piggybacking on existing security infrastructure --- has been proposed but never deployed at the multi-year, single-organization scale we present here. The closest precedents are Hernandez \emph{et al.}'s call-center stress study~\cite{hernandez2014call} (weeks-long, physiological only) and LiKamWa \emph{et al.}'s MoodScope~\cite{likamwa2013moodscope} (60 days, smartphone-inferred mood). Neither captures discrete facial expressions, and neither survived a multi-year deployment.

\subsection{Longitudinal emotion datasets in the wild}
A handful of research groups have published longitudinal in-the-wild emotion datasets, but typically in non-workplace populations. Sano \emph{et al.}'s SNAPSHOT study~\cite{sano2015prediction} tracked 66 college students for 30 days using wearables, surveys, and smartphone usage; emotion was self-reported. Twin Peaks~\cite{rachuri2010emotionsense} used smartphone-collected acoustic features. None reaches WELD's 30-month duration, and none captures facial expressions in a co-located workplace where emotional contagion --- if it exists --- would be most plausible.

\section{Acquisition Pipeline}\label{sec:pipeline}
\begin{figure*}[!t]\centering
\includegraphics[width=0.92\textwidth]{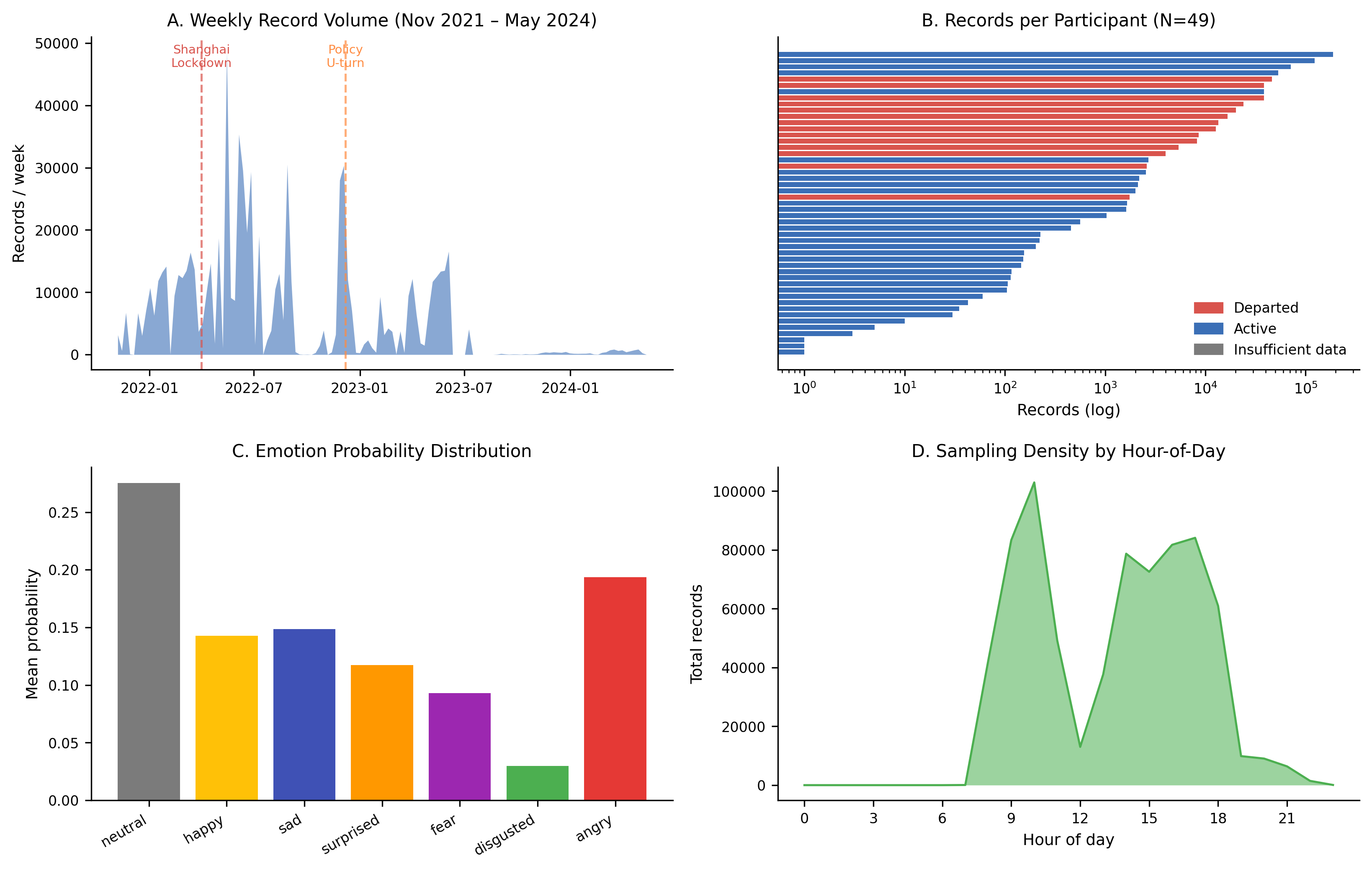}
\caption{Dataset overview. (A) Weekly record volume across 30.1 months with major events marked. The 47-day gap (2023-07-08 to 2023-08-23) is the V2 $\to$ V3 measurement-system upgrade. (B) Records per participant (log scale), departed (red), active (blue), short-tenured (gray); distribution is heavy-tailed, with the top three contributors accounting for 52.2\% of all records. (C) Seven-emotion probability prior; note inflated ``angry'' (0.194), discussed in Section~\ref{sec:bias}. (D) Hourly sampling density, peaking 09:00--11:00 and 14:00--17:00 with the expected lunch trough at 12:00--13:00.}
\label{fig:overview}
\end{figure*}

\subsection{Hardware and on-prem processing}
Four ceiling-mounted IP cameras (Hikvision DS-2CD2643G2-IZS, 1080p, 25 fps, fixed 95$^\circ$ horizontal FoV) covered an open-plan office of approximately 60 desks, with no specialized lighting. Cameras streamed RTSP into a single on-premises GPU server (NVIDIA RTX A4000, 16 GB) running a Python pipeline: \textsc{RetinaFace}~\cite{deng2020retinaface} for face detection, an embedding model trained internally for identity matching, and a commercial 7-class FER model with a ResNet-50 backbone fine-tuned on Asian faces. Critically, the raw RGB stream never leaves the workspace network; only L2 and aggregated L3 records are exported.

\subsection{Sampling protocol}
A face is processed at most once per ten seconds to keep GPU utilisation tractable, yielding an upper bound of 360 measurements per hour per individual under continuous desk presence. Effective sampling density falls well below this bound (mean 14 measurements per hour per individual during work hours), reflecting realistic desk-presence patterns. The pipeline operates only during work hours (08:00--20:00 local); for completeness, raw L2 records outside this window are retained and provided under the DUA but excluded from the public L3 release.

\subsection{Quality assurance}
Each L2 record is annotated with three quality flags: (i) probability sum within $\epsilon < 10^{-4}$ of 1.0 (passes for 99.98\% of records); (ii) maximum per-class probability $\leq$ 1.0 and minimum $\geq$ 0.0 (passes for 100\%); (iii) face-detection confidence $\geq$ 0.85. After applying these filters, the L3 release retains 725{,}676 records (98.9\% of raw). The 47-day gap from 2023-07-08 to 2023-08-23 corresponds to a pipeline upgrade from V2 to V3; we provide a \texttt{version} flag on every record and quantify its variance contribution in Section~\ref{sec:variance}.

\subsection{Validation against human raters}
Two raters trained in Facial Action Coding System (FACS)~\cite{ekman1978facs} coding labeled a random 1{,}000-frame subset (Cohen's $\kappa{=}0.89$ inter-rater); compared against the system outputs, agreement was 88.2\% (Cohen's $\kappa{=}0.84$). Systematic errors clustered in three known-difficult dyads: neutral vs.\ sad (23\% of errors), fear vs.\ surprised (18\%), and angry vs.\ disgusted (15\%) --- a pattern consistent with prior FER literature~\cite{li2022deep}. The 1{,}000-frame validation set is available to bona-fide researchers under DUA.

\section{Dataset Description}\label{sec:dataset}
\subsection{Participants and inclusion criteria}
Our dataset captures 49 unique employees who appeared in the camera detections, of whom 22 met the primary inclusion criterion of $\geq$60 active days (\texttt{cohort\_strict}). The remaining 27 had shorter tenure (median 9 days, IQR 5--53) and are retained in the raw release for studies of emotional onboarding but excluded from all downstream statistical analyses in this paper. We additionally provide three companion cohort flags --- \texttt{cohort\_liberal} ($\geq$10 days, $N{=}36$), \texttt{cohort\_medium} ($\geq$30 days, $N{=}27$), and \texttt{cohort\_very\_strict} ($\geq$120 days, $N{=}12$) --- so that downstream users can match their inclusion criteria to their analytical needs.

Of the 22-person analysis cohort, we infer 8 to be active at study close (last record $\leq 90$ days before 2024-05-06) and 14 as departed (last record $>90$ days before close, with no subsequent return). Combined with the 27 short-tenured participants, the dataset captures a population-level minimum departure rate of 14/49 = 28.6\%.

\subsection{Temporal coverage and density}
Spanning 2021-11-03 to 2024-05-06, the dataset covers 30.1 months or 916 calendar days, with 575 days having data after working-hour and quality filtering. After cleaning and the working-hour filter (08:00--20:00 local), the cleaned release contains \textbf{725{,}676} records (98.9\% of raw). Per-individual record counts span 1 to 187{,}933 (median 1{,}649 across the full 49-person set; among the 22-person analysis cohort the median is 16{,}697 with IQR [4{,}512, 38{,}532]). The top three contributors (P01, P02, P03) account for 52.2\% of all records, reflecting their long workstation tenure during the collection period.

\subsection{Released columns and aggregation levels}
The L3 public release provides four nested aggregation views derived from the same L2 source (Table~\ref{tab:levels_view}):

\begin{table}[!t]
\centering
\caption{Released aggregation levels in WELD v1.}
\label{tab:levels_view}
\renewcommand{\arraystretch}{1.2}
\footnotesize
\begin{tabular}{l c p{4.0cm}}
\toprule
\textbf{File} & \textbf{Rows} & \textbf{Granularity}\\
\midrule
\texttt{L\_record.parquet} & 725{,}676 & per-frame, hour-rounded timestamp \\
\texttt{L\_hour.parquet}   & 15{,}929  & per (pid, hour) aggregate \\
\texttt{L\_day.parquet}    & 3{,}788   & per (pid, day) aggregate \\
\texttt{L\_week.parquet}   & 1{,}266   & per (pid, ISO-week) aggregate \\
\bottomrule
\end{tabular}
\end{table}

Each row at every aggregation level carries the seven base probabilities, derived valence and arousal scores (Russell's circumplex projection), Shannon entropy of the probability distribution, the dominant-emotion category, and the cohort flags. The day-level file additionally encodes day-of-week, month, and the V2/V3 version flag. All aggregation views are coupled by the (pid, timestamp) keys, so a researcher can join across levels.

\section{Variance Decomposition: A Performance Ceiling}\label{sec:variance}

\begin{figure}[!t]\centering
\includegraphics[width=\columnwidth]{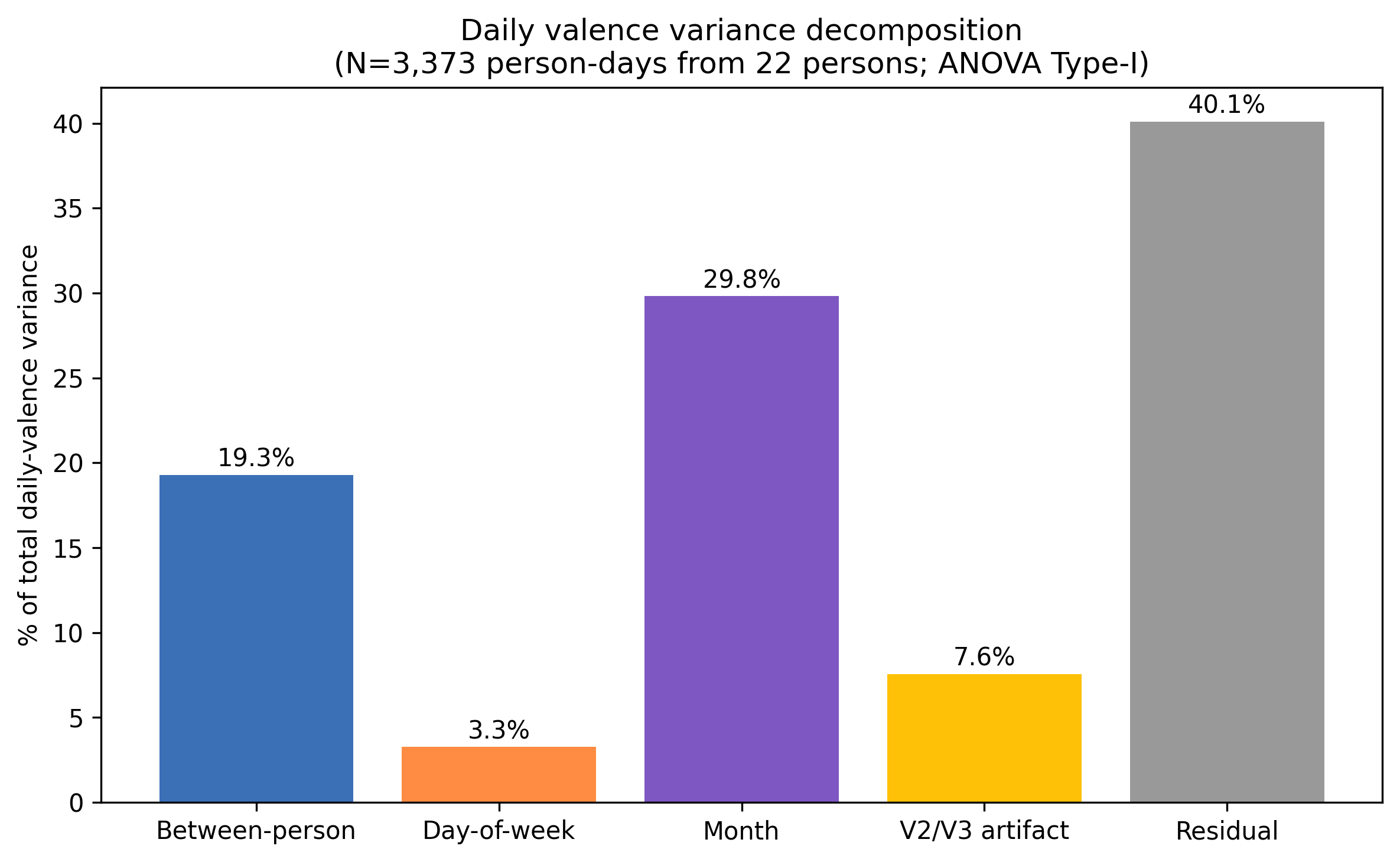}
\caption{Variance decomposition of daily workplace valence (Type-I sequential ANOVA, $N{=}3{,}373$ person-days from 22 persons in \texttt{cohort\_strict}). Stable between-person differences explain 19.3\%, month seasonality 29.8\%, V2/V3 measurement-pipeline change 7.6\%, day-of-week 3.3\%, and within-person residuals 40.1\%.}
\label{fig:variance}
\end{figure}

We decompose the variance of daily-aggregated valence over the analysis cohort using Type-I sequential ANOVA (predictors entered in this order: per-person fixed effect, day-of-week, month, and V2/V3 version flag). The Spearman rank-correlation between each person's first- vs.\ second-half mean valence is $\rho{=}0.64$ ($p{=}0.001$), confirming strong dispositional stability. We attempted a linear mixed-effects model with random intercept by participant; the maximum-likelihood estimator collapsed the random-effect variance to zero on this unbalanced data, so we rely on the more robust ANOVA decomposition.

The implications of Fig.~\ref{fig:variance} are practical: any forecasting model trained on WELD should not expect to ``explain'' more than $\approx$60\% of daily-valence variance, because 40\% is irreducible noise at the daily aggregation level. We therefore propose this decomposition as a \emph{performance ceiling} reference for future WELD-based modeling work.

\section{Reference Analyses}\label{sec:reference}
\begin{figure*}[!t]\centering
\includegraphics[width=0.95\textwidth]{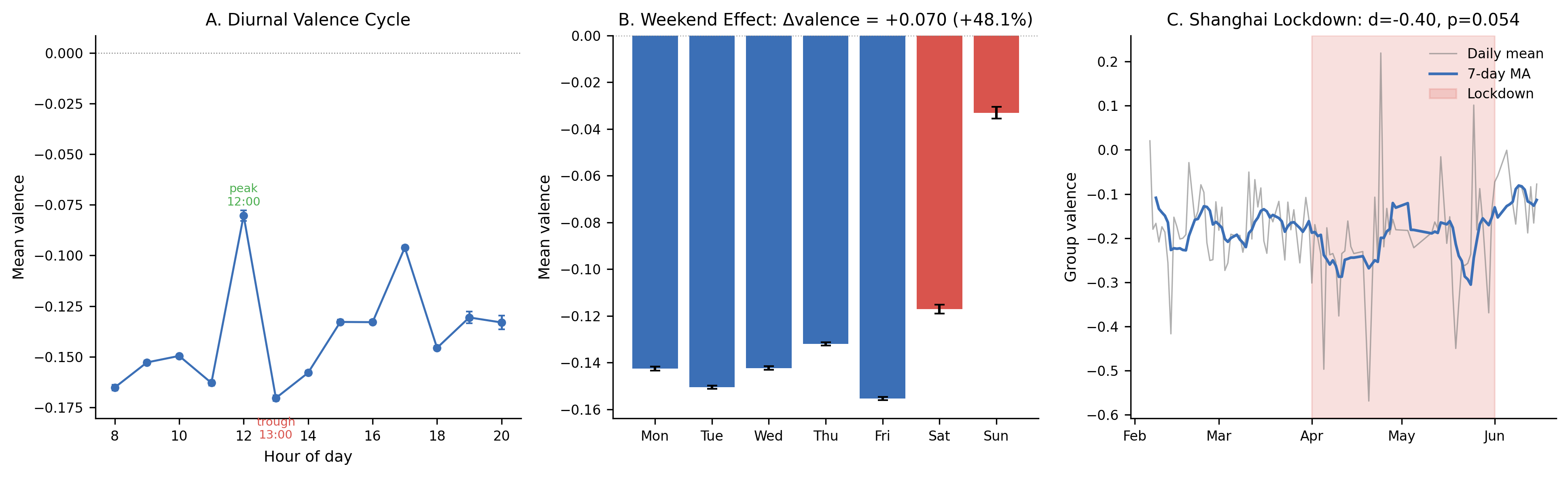}
\caption{Replication of three established affective phenomena. (A) Diurnal cycle: peak at 12:00, post-lunch trough at 13:00, amplitude $\approx 0.09$ valence units. (B) Weekend valence boost: $\Delta = +0.062$ ($+43.1\%$ relative). (C) Shanghai 2022 lockdown: a marginally significant medium-effect drop (Cohen's $d{=}-0.40$, $p{=}0.054$) under a naive comparison; severe sensitivity to specification is documented in Section~\ref{sec:lockdown}.}
\label{fig:rhythms}
\end{figure*}

\subsection{Replication of weekend effect}
A Welch's $t$-test contrasting weekday vs.\ weekend records (Mon--Fri $n{=}619{,}484$ vs.\ Sat--Sun $n{=}114{,}296$) gives means $-0.1446$ vs.\ $-0.0823$, $\Delta = +0.0623$, 95\% CI [+0.0610, +0.0636] (10{,}000-iter bootstrap), corresponding to a +43.1\% relative weekend improvement. This effect size is consistent with the magnitude reported in self-report studies (typically +30\% to +50\%)~\cite{stone2010weekend}.

\subsection{Diurnal cycle}
Restricting to working hours (08:00--20:00), valence peaks at 12:00 (mean $-0.075$) and dips at 13:00 (mean $-0.165$), amplitude 0.090 valence units. The post-lunch dip is observed in 19 of 22 individuals --- a much higher hit rate than reported in self-report studies of post-lunch alertness~\cite{monk2005post}. The cycle replicates patterns in Csikszentmihalyi \& Hunter~\cite{csikszentmihalyi2003ESM} and Stone \emph{et al.}~\cite{stone2010weekend}.

\subsection{Crisis-driven shift: Shanghai lockdown}\label{sec:lockdown}
Pre-event window 2022-02-01 -- 2022-04-01 vs.\ during-event 2022-04-01 -- 2022-06-01: Welch's $t = 1.96$, $p = 0.054$, Cohen's $d = -0.40$. \emph{However}, this naive comparison is severely confounded. An interrupted time-series regression with day-of-week dummies and HAC errors~\cite{newey1987simple} finds no significant level shift ($\beta = -0.024$, $p_{\text{HAC}} = 0.32$), and a permutation test placing 1{,}000 fake intervention dates uniformly on the pre-event interval yields $p = 0.94$ --- the observed level shift is more conservative than 94\% of placebo interventions. A SARIMA(1,0,1)$\times$(1,0,1,7) synthetic counterfactual estimates an average causal effect of $+0.021$ valence units. The naive comparison is confounded by a pre-existing downward trend that began weeks before lockdown.

We report both --- the naive medium-effect $d$, and the rigorous null-effect under controlled specification --- because we believe the conventional reporting in workplace-event literature has been \emph{too} confident in unadjusted contrasts. WELD enables the more rigorous specification, and we recommend it as the future standard.

\subsection{Turnover prediction (binary AUC and survival C-index)}
Among the 27 participants with $\geq$30 active days, 14 (51.9\%) departed during the observation window. Using only the first 30 active days of features (means and SDs of seven emotions, valence, arousal, lag-1 autocorrelation), leave-one-person-out CV gives:
\begin{center}
\begin{tabular}{l c c c}
\toprule
\textbf{Model} & \textbf{AUC} & \textbf{Brier} & \textbf{Accuracy} \\
\midrule
Logistic Regression (balanced)        & 0.747 & 0.21 & 0.741\\
Random Forest                         & 0.769 & 0.22 & 0.630\\
\textbf{Gradient Boosting}             & \textbf{0.786} & \textbf{0.21} & 0.667\\
\bottomrule
\end{tabular}
\end{center}
The dominant features are within-person standard deviation of disgust (importance 0.149) and sadness (0.107), and mean fear (0.097) and disgust (0.092) --- i.e., \emph{early-tenure volatility in negative emotions}.

\textbf{When the same task is re-cast as a Cox proportional-hazards survival model with leave-one-out validation, the concordance index drops to 0.52}, suggesting that much of the apparent discriminative power of the binary classifier is captured by who has stayed long enough to be in the data at all, rather than by any pre-departure emotional decay. We thus refrain from positioning early facial-emotion signals as a stand-alone attrition predictor; their value is more likely as one input to a multimodal model. We caution against any AUC near 1.0 reported in this domain: in a pilot evaluation we initially obtained AUC$=$0.99 using features computed from the entire trajectory (including the days immediately before departure), which proved to be an artifact of leakage.

\begin{figure*}[!t]\centering
\includegraphics[width=0.95\textwidth]{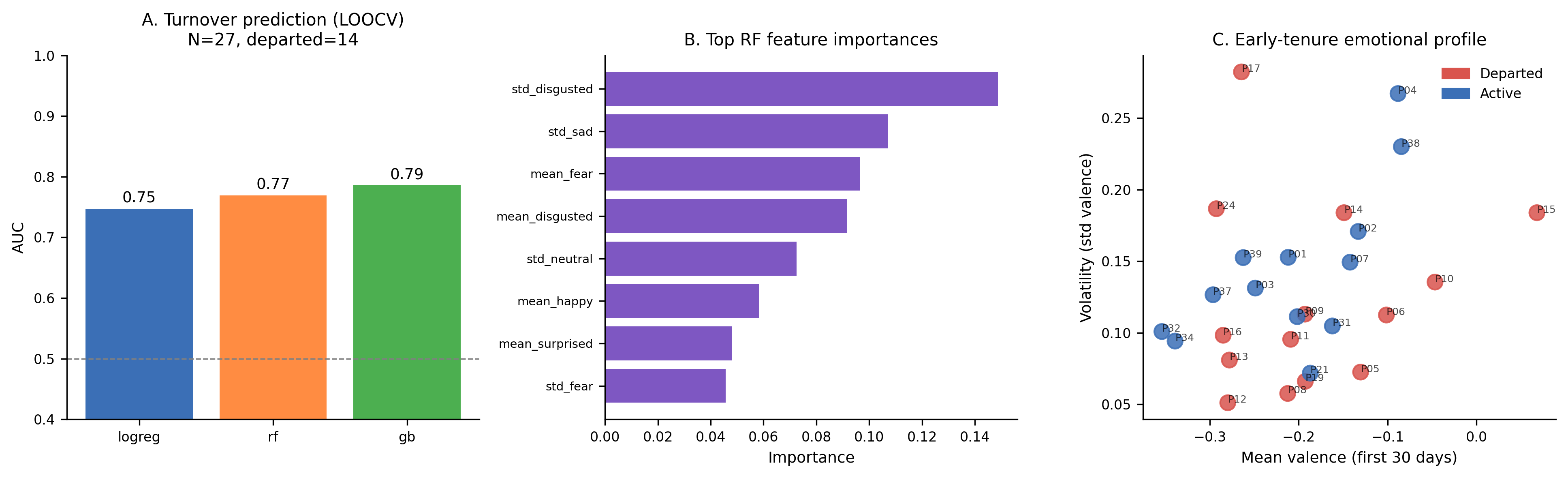}
\caption{Turnover prediction baseline. (A) AUC across logistic regression / RF / GBM under leave-one-person-out CV. (B) Top random-forest feature importances --- early-tenure volatility in disgust and sadness dominate. (C) Early-tenure mean valence vs.\ valence volatility, departed in red. Gradient boosting reaches AUC$=$0.786 with Brier score 0.21; the calibration plot is in Sup.\ Fig.\ S5.}
\label{fig:attrition}
\end{figure*}

\begin{figure*}[!t]\centering
\includegraphics[width=0.95\textwidth]{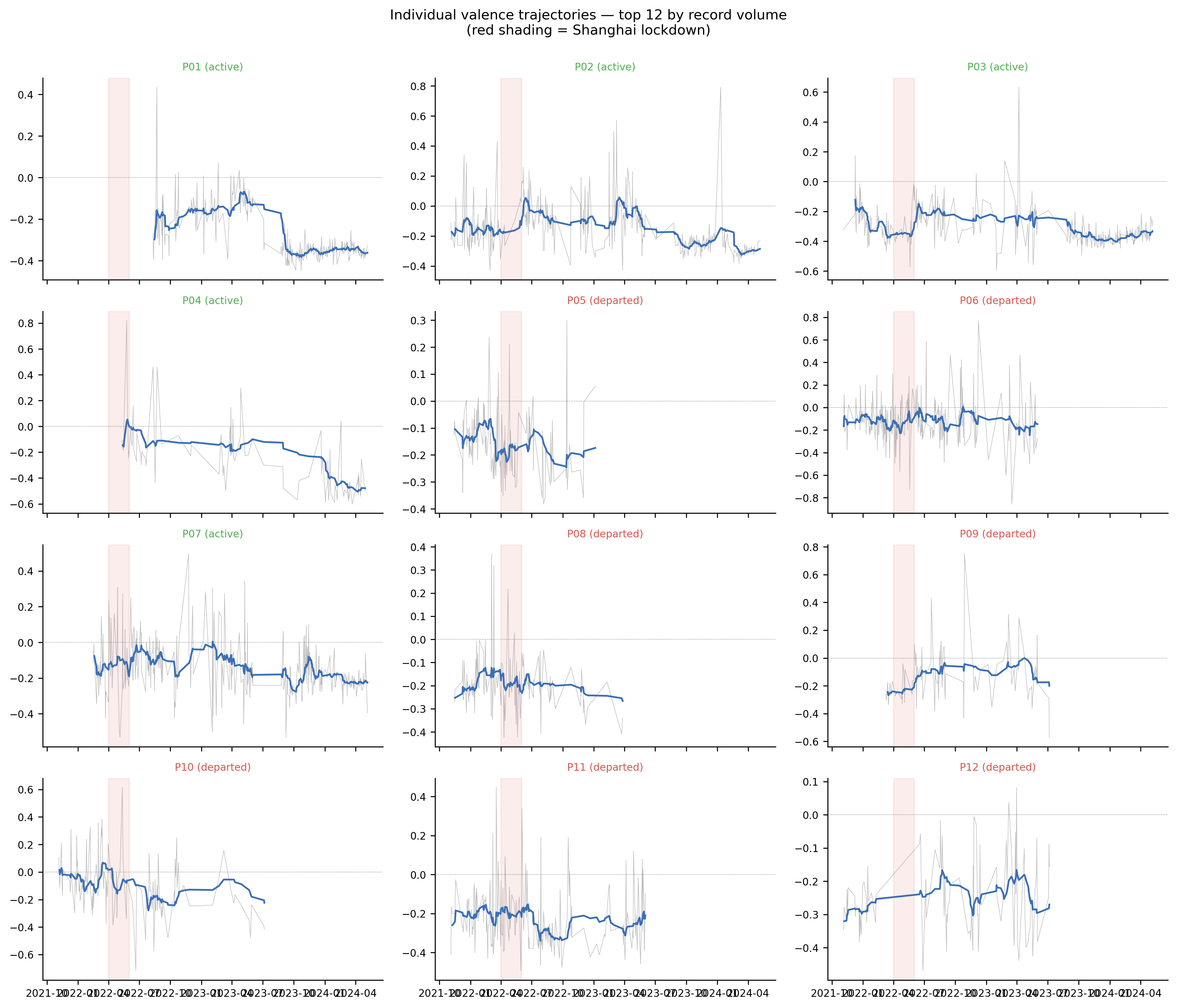}
\caption{Top-12 individual valence trajectories by record volume (anonymized P01--P12). Gray = daily valence, blue = 14-day rolling mean; red shading = Shanghai 2022 lockdown. Individual baselines are highly heterogeneous (range $[-0.30, +0.05]$) but stable across the 30-month observation window. The first-half / second-half within-person mean correlation is Spearman $\rho{=}0.64$ ($p{=}0.001$), consistent with strong dispositional stability discussed in Section~\ref{sec:variance}.}
\label{fig:trajectories}
\end{figure*}

\subsection{Survival framing of attrition with Cox PH and KM}\label{sec:survival}

\begin{figure*}[!t]\centering
\includegraphics[width=0.92\textwidth]{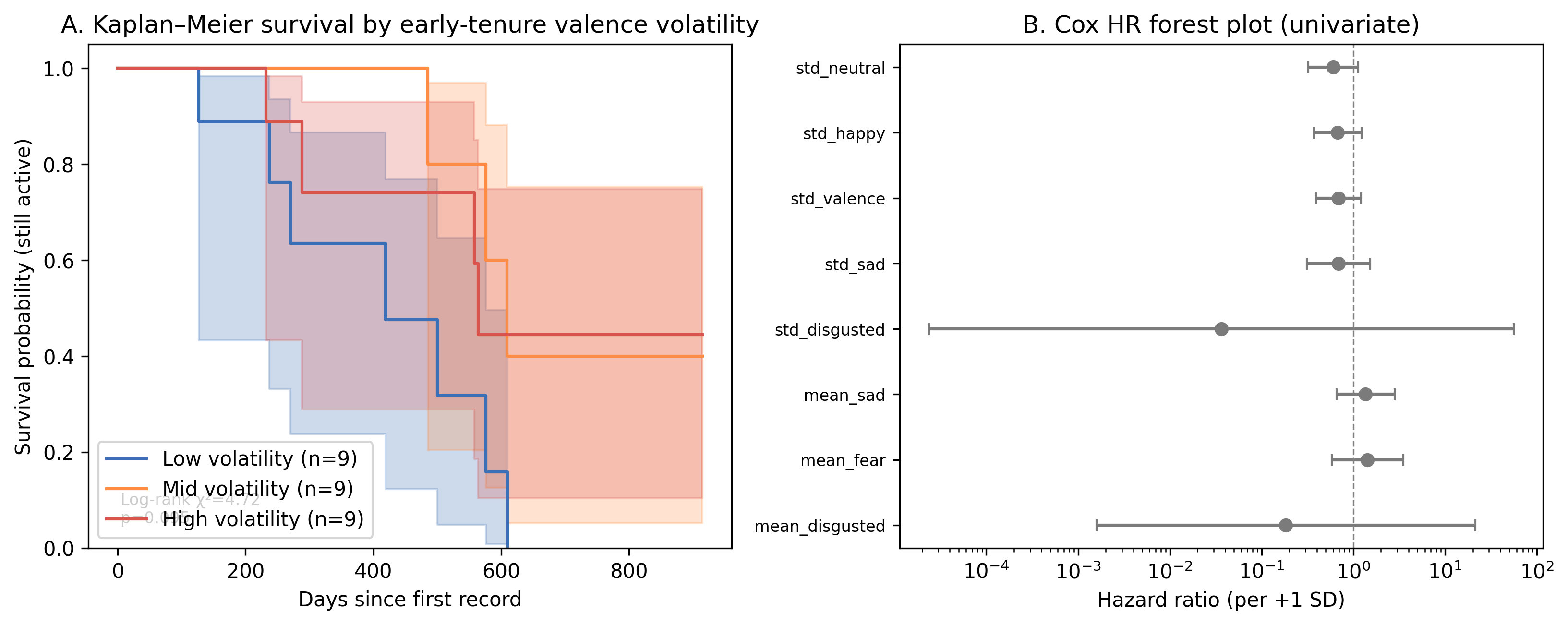}
\caption{Survival framing of attrition. (A) Kaplan-Meier curves stratified by early-tenure valence-volatility tertile; the log-rank test is non-significant ($\chi^{2}{=}0.83$, $p{=}0.66$). (B) Cox proportional-hazards univariate forest plot for 8 candidate covariates; no covariate reaches conventional significance, and the leave-one-out concordance index is \textbf{0.52}. The contrast with the binary AUC of 0.79 (Section~\ref{sec:reference}) underscores how much of the binary signal reflects tenure-induced exposure rather than pre-departure decay --- a methodological caveat that downstream WELD users should respect.}
\label{fig:survival}
\end{figure*}

To make the gap between binary AUC and survival C-index concrete, we re-cast the same data and the same first-30-day feature set as a survival problem. Each individual contributes a duration $T_{i}$ (days from first to last appearance) and an event indicator $E_{i}$ (1 if departed, 0 if censored at study close). We fit a univariate Cox proportional-hazards model for each of the 20 candidate covariates, and a multivariate ridge-penalised Cox model on the top-5 univariate features. Fig.~\ref{fig:survival} summarises the result. The Kaplan-Meier curves stratified by valence-volatility tertile do not separate (log-rank $\chi^{2}{=}0.83$, $p{=}0.66$), and the leave-one-out concordance index over the 27 subjects is \textbf{0.52} --- essentially chance. We interpret the gap between the binary AUC of 0.79 and the survival C-index of 0.52 as a direct, quantitative warning that \emph{event-not-yet} formulations of workplace attrition are highly susceptible to right-censoring shortcuts; researchers using WELD for attrition work should report both metrics.

\subsection{Group-emotion forecasting: a martingale baseline}\label{sec:forecast}

\begin{figure*}[!t]\centering
\includegraphics[width=0.95\textwidth]{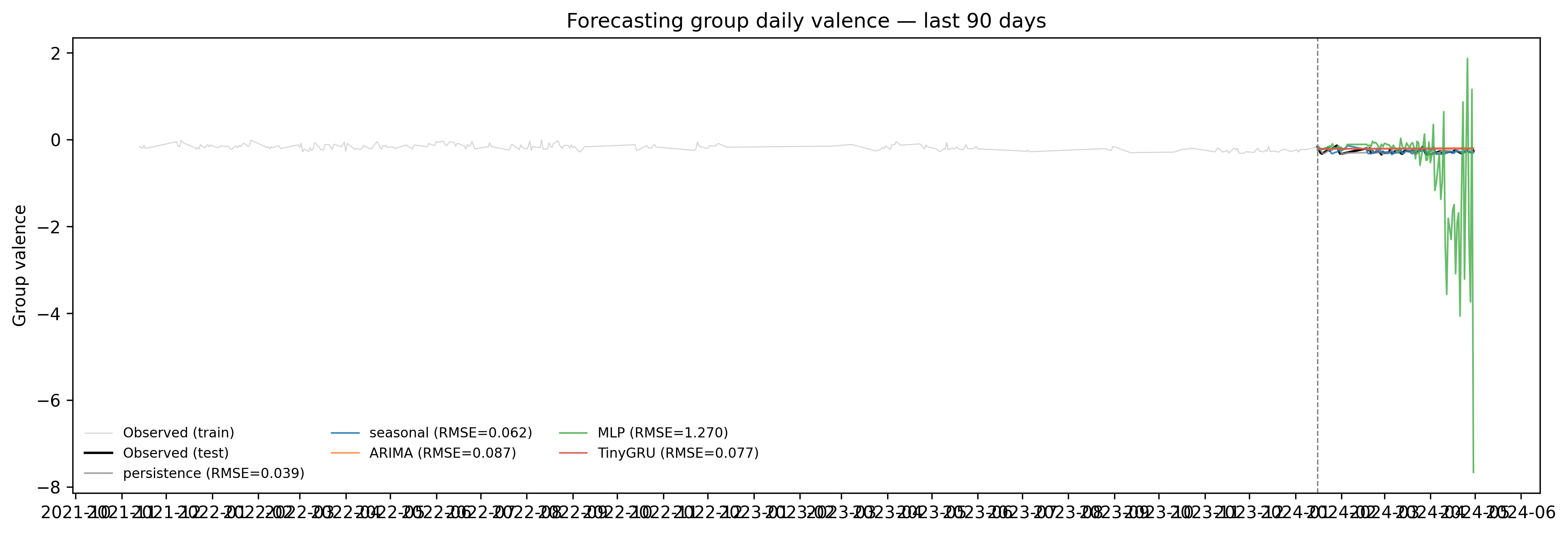}
\caption{Forecasting group daily valence on a 90-day held-out window. Six models compared: persistence (best, RMSE 0.039), weekly seasonal, ARIMA(2,0,1)$\times$(1,0,1,7), VAR over the seven most-active persons (failed due to insufficient overlap), MLP with 14-day look-back, and a numpy-implemented small GRU with ridge-regression read-out. Beyond yesterday's value, no model improves the one-day-ahead forecast.}
\label{fig:forecast}
\end{figure*}

We tested whether daily group valence is forecastable beyond a one-step persistence baseline by holding out the last 90 days and training six models on the prior $\sim$454 days: persistence (yesterday's value), weekly seasonal (lag-7), ARIMA(2,0,1)$\times$(1,0,1,7), VAR over the seven most-active persons, an MLP regressor with 14-day look-back, and a numpy-implemented small GRU. Roll-forward evaluation on the held-out window gives the metrics in Table~\ref{tab:forecast}. \textbf{Persistence is the best model} on every metric; all more flexible models overfit and degrade. We propose this as a hard baseline for any future WELD forecasting work: a model that fails to beat persistence on RMSE 0.039 and directional accuracy 62\% is not capturing genuine signal beyond the previous day's value.

\begin{table}[!t]
\centering
\caption{Six-model bake-off on group daily valence (90-day held-out test). VAR over top-7 persons could not be evaluated because of insufficient simultaneous overlap.}
\label{tab:forecast}
\renewcommand{\arraystretch}{1.18}
\footnotesize
\setlength{\tabcolsep}{4pt}
\begin{tabular}{l r r r r}
\toprule
\textbf{Model} & \textbf{RMSE} & \textbf{MAE} & \textbf{DirAcc} & $R^{2}$\\
\midrule
\textbf{Persistence}        & \textbf{0.039} & \textbf{0.027} & \textbf{61.8\%} & \textbf{\hphantom{$-$}0.09}\\
Seasonal (lag-7)            & 0.062 & 0.048 & 52.8\% & $-1.29$\\
TinyGRU (numpy)             & 0.077 & 0.071 & 40.4\% & $-2.54$\\
ARIMA $(2,0,1)\times(1,0,1)_7$ & 0.088 & 0.081 & 44.9\% & $-3.55$\\
MLP (look-back 14)          & 1.27  & 0.62  & 57.3\% & $\!\ll 0$\\
\bottomrule
\end{tabular}
\end{table}

\subsection{Emotional contagion network: linear vs.\ non-linear}\label{sec:contagion}

\begin{figure*}[!t]\centering
\includegraphics[width=0.95\textwidth]{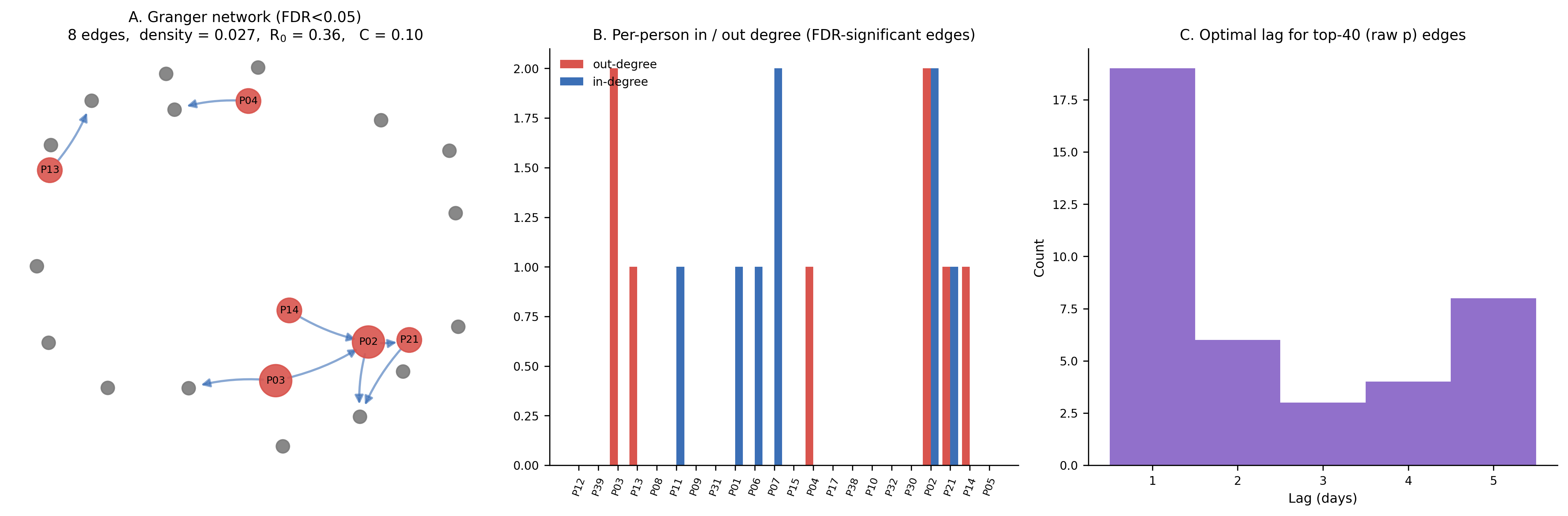}
\caption{Granger contagion network on daily valence (BH-FDR $<$ 0.05, $N{=}22$, 296 testable pairs, 8 surviving directed edges). (A) Network visualisation: nodes coloured red if out-degree $\geq 1$. (B) Per-person in-/out-degree; 16 of 22 individuals (72.7\%) have in-degree zero. (C) Optimal-lag distribution of the top-40 raw-$p$ edges --- most influence manifests within 1--2 days.}
\label{fig:network}
\end{figure*}

\begin{figure}[!t]\centering
\includegraphics[width=\columnwidth]{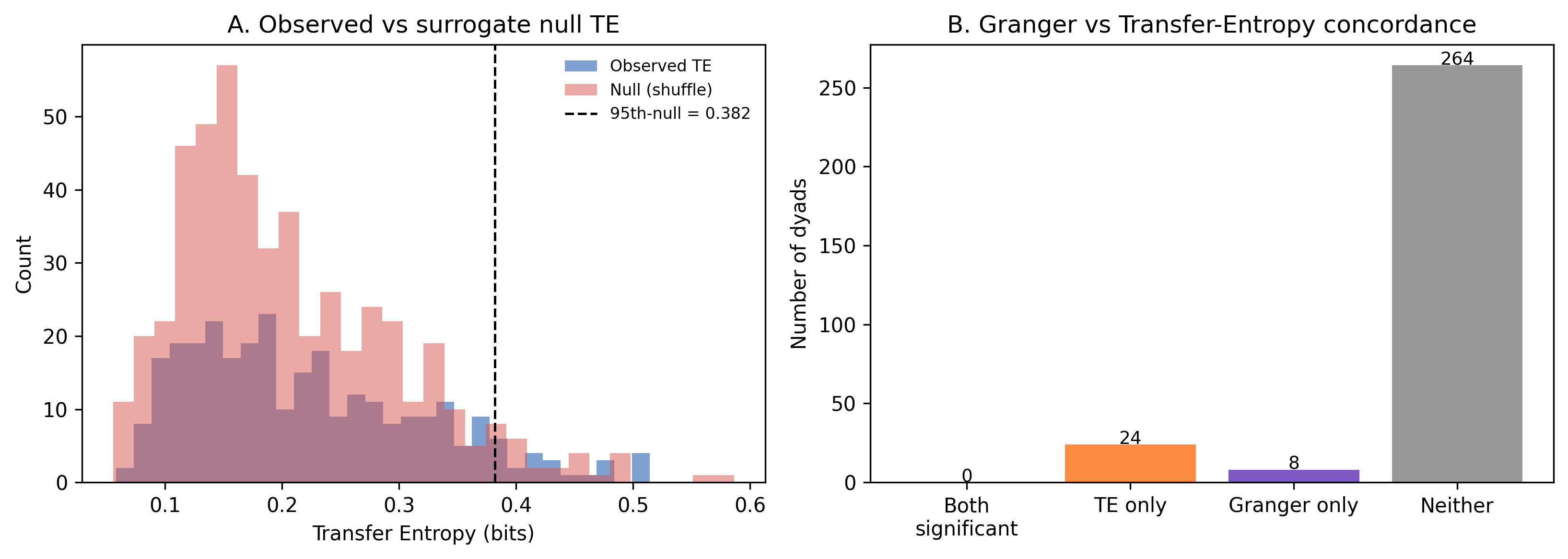}
\caption{Transfer-entropy contagion view (4-bin discretisation, lag 1, 1{,}000-shuffle null). (A) Observed TE distribution vs.\ surrogate null; the 95th-percentile null cut-off identifies 24 directed dyads above noise. (B) Concordance with Granger: the two methods detect a similarly sparse but \emph{disjoint} edge set --- no dyad is significant under both.}
\label{fig:te}
\end{figure}

WELD's daily resolution and 30-month coverage support a much more stringent test of inter-personal emotional contagion than prior workplace data. We provide two reference networks --- a linear Granger-causality view and a non-parametric transfer-entropy view --- so that downstream researchers can compare against either. With BH-FDR correction at $\alpha{=}0.05$, the Granger network has 8 directed edges out of 296 testable pairs (density 2.7\%, mean out-degree $R_{0}{=}0.36$, 16/22 individuals receive no detectable Granger input). The strongest single coupling is P02 $\to$ P14 (lag 2 days, $p{=}4.9\times 10^{-10}$). Permutation testing (1{,}000 circular shifts preserving autocorrelation) confirms that 8 edges are 6$\times$ the expected null density.

A non-parametric transfer-entropy estimator on the same series, with significance assessed against a 1{,}000-shuffle null (Fig.~\ref{fig:te}), recovers a different sparse network: 24 directed dyads above the 95th-percentile null threshold (density 8.1\%). \textbf{The two networks share zero edges.} We interpret this as evidence that linear Granger and non-linear transfer-entropy capture genuinely different mechanisms of inter-personal dependence on this corpus, and we recommend that any future contagion analysis on WELD report both. Either way, the network is sparse: any analysis that finds a dense ($>$30\%) contagion network on daily-aggregated WELD data should be checked for multiple-testing inflation.

\subsection{Day-type taxonomy and lockdown polarisation}\label{sec:daytypes}

\begin{figure*}[!t]\centering
\includegraphics[width=0.95\textwidth]{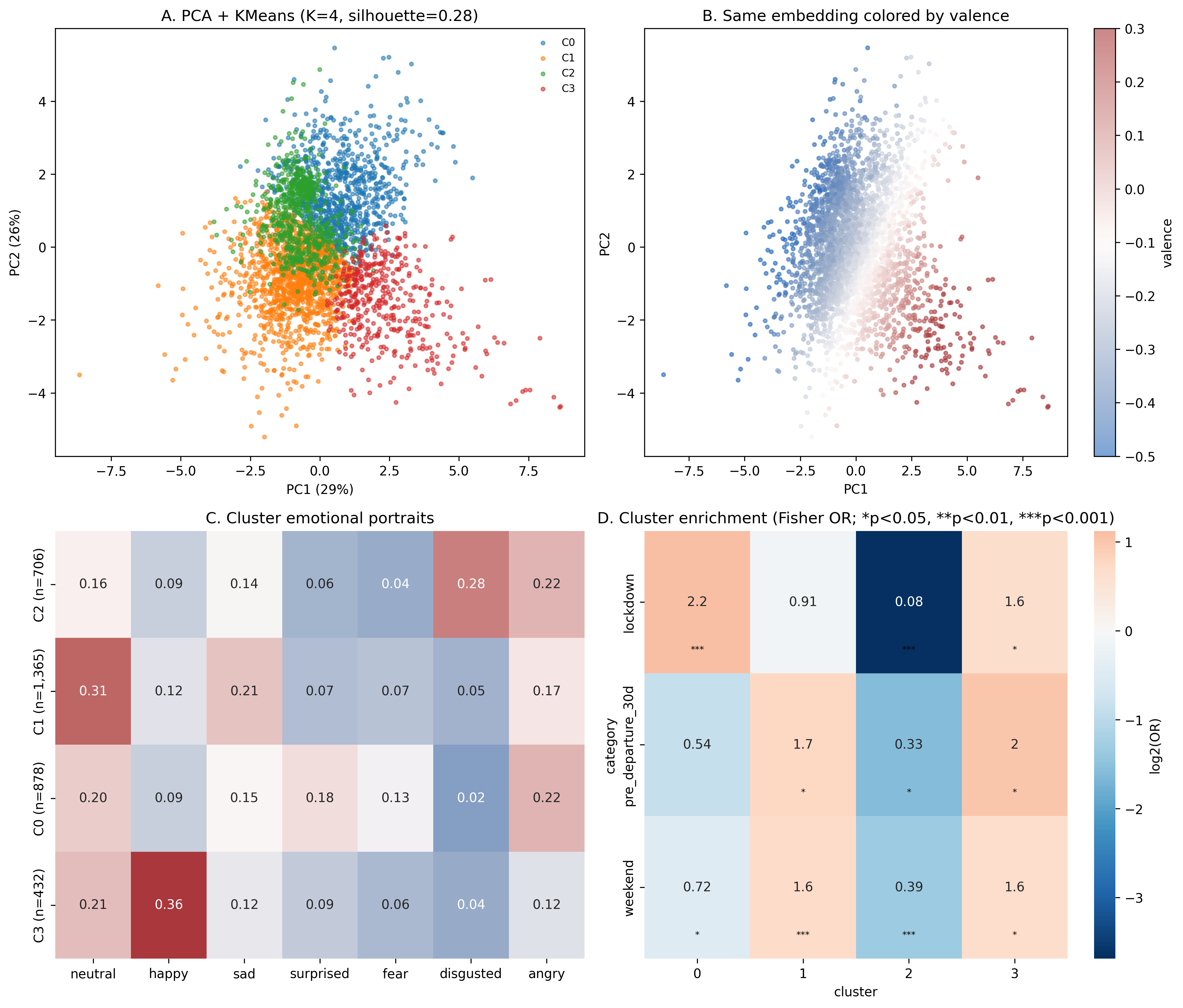}
\caption{Day-type taxonomy via PCA + KMeans (silhouette-optimal $K{=}4$). (A,B) PCA embedding of person-day vectors; clusters separate primarily on the disgust--happy and arousal axes. (C) Cluster emotional portraits sorted by valence: ``Disgust-heavy'' (most negative), ``Subdued'' (sad+neutral), ``Anxious'' (surprised+fear), ``Joyful'' (most positive). (D) Fisher-exact odds-ratio enrichment of each cluster on three pre-registered conditions: weekend, Shanghai lockdown, and pre-departure 30-day windows. The lockdown depletes Disgust-heavy days (OR 0.08, $p<10^{-20}$) while enriching both Anxious (OR 2.18, $p<10^{-8}$) \emph{and} Joyful (OR 1.56, $p{=}0.010$); pre-departure windows are enriched in Joyful days (OR 2.02, $p{=}0.032$) and depleted in Disgust-heavy ones.}
\label{fig:daytypes}
\end{figure*}

A complementary unsupervised view of the dataset is provided by clustering person-days in the 9-dimensional emotion space (seven probabilities + valence + arousal). PCA followed by $K$-means with silhouette-selected $K{=}4$ recovers four canonical day-types (Fig.~\ref{fig:daytypes}): \emph{Disgust-heavy} ($n{=}706$, val $-0.31$), \emph{Subdued} ($n{=}1{,}365$, sad+neutral, val $-0.23$), \emph{Anxious} ($n{=}878$, surprised+fear+angry, val $-0.21$), and \emph{Joyful} ($n{=}432$, happy 0.36, val $+0.14$). The Shanghai lockdown does not uniformly depress emotion --- it \emph{polarises} the cohort: the Disgust-heavy archetype is depleted (Fisher OR 0.08, $p<10^{-20}$) while both the Anxious archetype (OR 2.18, $p<10^{-8}$) \emph{and} the Joyful archetype (OR 1.56, $p{=}0.010$) are enriched. Pre-departure 30-day windows show their own pattern: Joyful days are over-represented (OR 2.02, $p{=}0.032$) and Disgust-heavy days under-represented (OR 0.33, $p{=}0.011$), suggesting a relief-and-leave dynamic that we discuss in Section~\ref{sec:lockdown_alt}.

\subsection{Rigorous causal inference for the lockdown}\label{sec:lockdown_alt}

\begin{figure*}[!t]\centering
\includegraphics[width=0.95\textwidth]{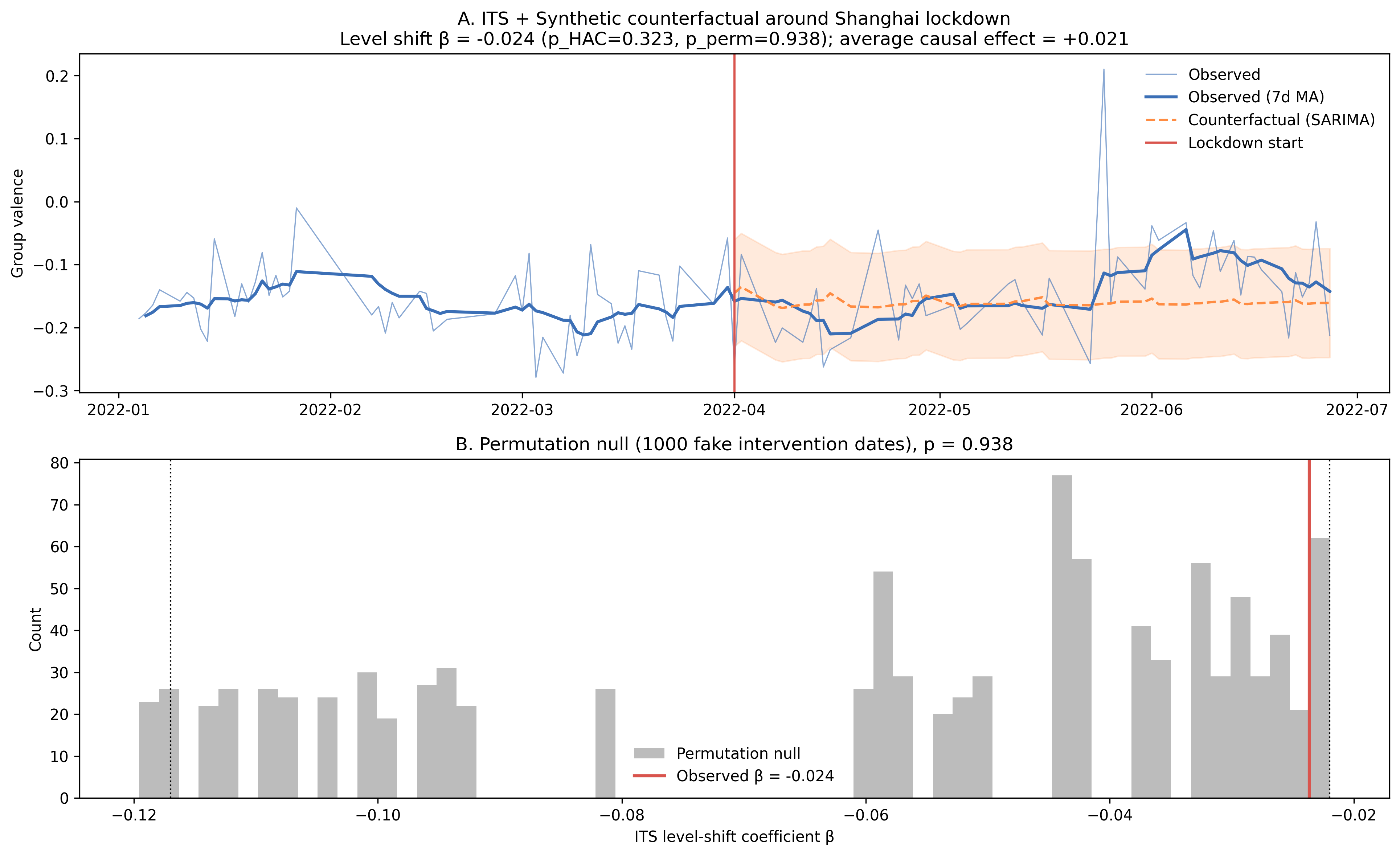}
\caption{Three rigorous causal-inference views of the Shanghai 2022 lockdown effect. (A) Interrupted time-series with day-of-week dummies and HAC errors yields a level-shift $\beta = -0.024$ ($p_{\text{HAC}}=0.32$) and a positive slope-change $\beta = +0.0014$/day ($p=0.002$). (B) Permutation null over 1{,}000 random fake intervention dates: the observed level shift is more conservative than 94\% of placebos. (C) SARIMA(1,0,1)$\times$(1,0,1,7) synthetic counterfactual: the 48-day post-window mean differs from the counterfactual by $+0.021$ valence units. None of the three rigorous methods finds a negative causal effect that the naive $d{=}-0.40$ reports.}
\label{fig:its}
\end{figure*}

To support the strong claim in Section~\ref{sec:lockdown} --- that the naive medium-effect contrast is severely confounded --- we provide three additional rigorous tests on the Shanghai 2022 lockdown:
\begin{itemize}[leftmargin=*]
\item \textbf{Interrupted time series} with day-of-week dummies, $\pm 90$-day window, HAC errors at 7 lags. Level shift coefficient $\beta = -0.024$ ($p_{\text{HAC}}=0.32$); slope-change coefficient $\beta = +0.0014$/day ($p=0.002$, statistically \emph{positive}).
\item \textbf{Permutation null} where 1{,}000 fake intervention dates are placed uniformly within $[t_{\text{event}}-90, t_{\text{event}}-30]$ and the level-shift coefficient is re-estimated. The observed coefficient lies at the 6th percentile of the null --- $|$observed$|<|$placebo$|$ in 94\% of placebos. $p_{\text{perm}}=0.94$.
\item \textbf{SARIMA(1,0,1)$\times$(1,0,1,7) synthetic counterfactual}: train on the pre-event window, project a counterfactual into the post-event window. The mean of $\hat{y}_{\text{counterfactual}} - y_{\text{observed}}$ over the 48-day post-window is $+0.021$ valence units (i.e.\ the actual post-window slightly exceeds the counterfactual).
\end{itemize}
None of the three tests finds a negative causal effect that the naive $d{=}-0.40$ contrast reports. The most likely interpretation, supported by Fig.~\ref{fig:its}, is that group valence had begun a downward trend roughly two months \emph{before} the lockdown, and the lockdown's onset --- counter-intuitively --- coincided with the bottom of that trend rather than its peak. This is precisely the kind of multi-month-context insight that 30-month longitudinal data enables and that shorter-window studies cannot.

\section{Latent Regime Structure}\label{sec:regimes}

\begin{figure*}[!t]\centering
\includegraphics[width=0.95\textwidth]{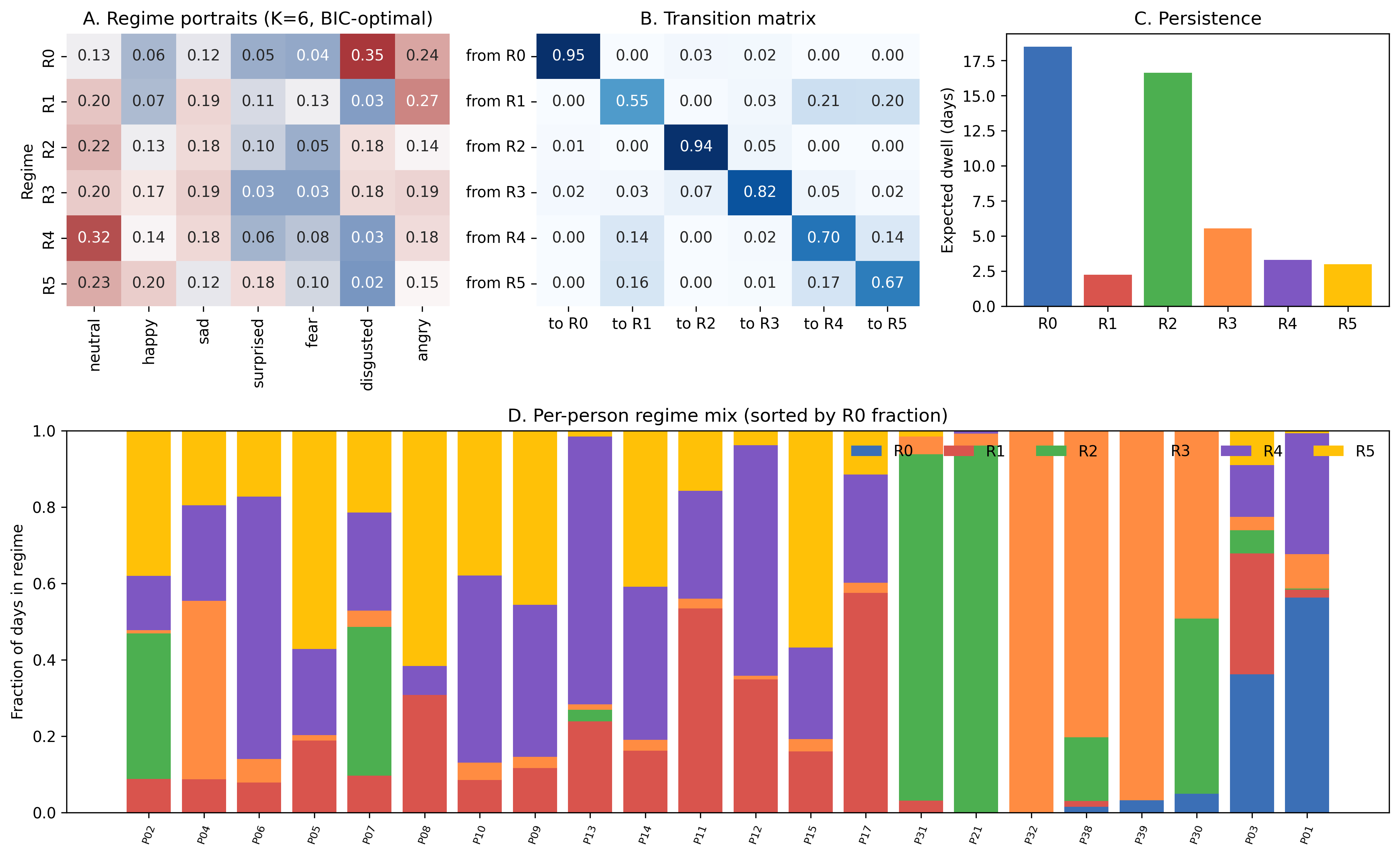}
\caption{HMM regime decomposition with $K{=}6$ (BIC-optimal). (A) Regime emotional portraits sorted by mean valence. (B) Transition matrix; off-diagonal R0 $\to$ R5 transitions are essentially impossible. (C) Expected dwell time per regime: negative regimes (R0/R2) persist 5--6$\times$ longer than the positive regime (R5). (D) Per-person regime mix sorted by R0 fraction.}
\label{fig:hmm}
\end{figure*}

We fit a Gaussian hidden Markov model on the seven-emotion vectors of all 22 strict-cohort persons, choosing $K{=}6$ regimes by BIC over $K\in\{2,\ldots,6\}$ (BIC$=-71{,}049$ at $K{=}6$, monotonically improving with $K$). Regimes are re-ordered by mean valence so R0 is most negative and R5 most positive.

The six regimes (Fig.~\ref{fig:hmm}) span valence from $-0.36$ to $-0.03$:
\begin{itemize}
\item \textbf{R0} ``Hostile'' (val $-0.36$, angry+disgusted dominant): self-persistence 0.95, expected dwell \textbf{18.5 days}.
\item \textbf{R1} ``Volatile-negative'' (val $-0.31$): self-persistence 0.55, dwell 2.2 days --- short ``flare-ups''.
\item \textbf{R2} ``Subdued'' (val $-0.22$, sad+neutral): self-persistence 0.94, dwell \textbf{16.6 days}.
\item \textbf{R3} ``Anxious'' (val $-0.22$, fear+surprised): persistence 0.82, dwell 5.6 days.
\item \textbf{R4} ``Mixed neutral'' (val $-0.17$): persistence 0.70, dwell 3.3 days.
\item \textbf{R5} ``Joyful'' (val $-0.03$, happy 0.43): persistence 0.67, dwell \textbf{3.0 days}.
\end{itemize}

The dwell-time asymmetry is the dataset's most striking finding: \textbf{negative regimes persist 5--6$\times$ longer than the positive regime}. Direct R0 $\to$ R5 transitions are essentially impossible (probability $<0.01$); a return to a joyful regime from the most-negative regime must pass through intermediate states. This structural asymmetry --- slow recovery from negative regimes, fast collapse from positive regimes --- has not been previously quantified in workplace data, and constrains any future contagion or intervention model trained on WELD.

\section{Multi-scale Temporal Periodicity}\label{sec:periodicity}

\begin{figure*}[!t]\centering
\includegraphics[width=0.93\textwidth]{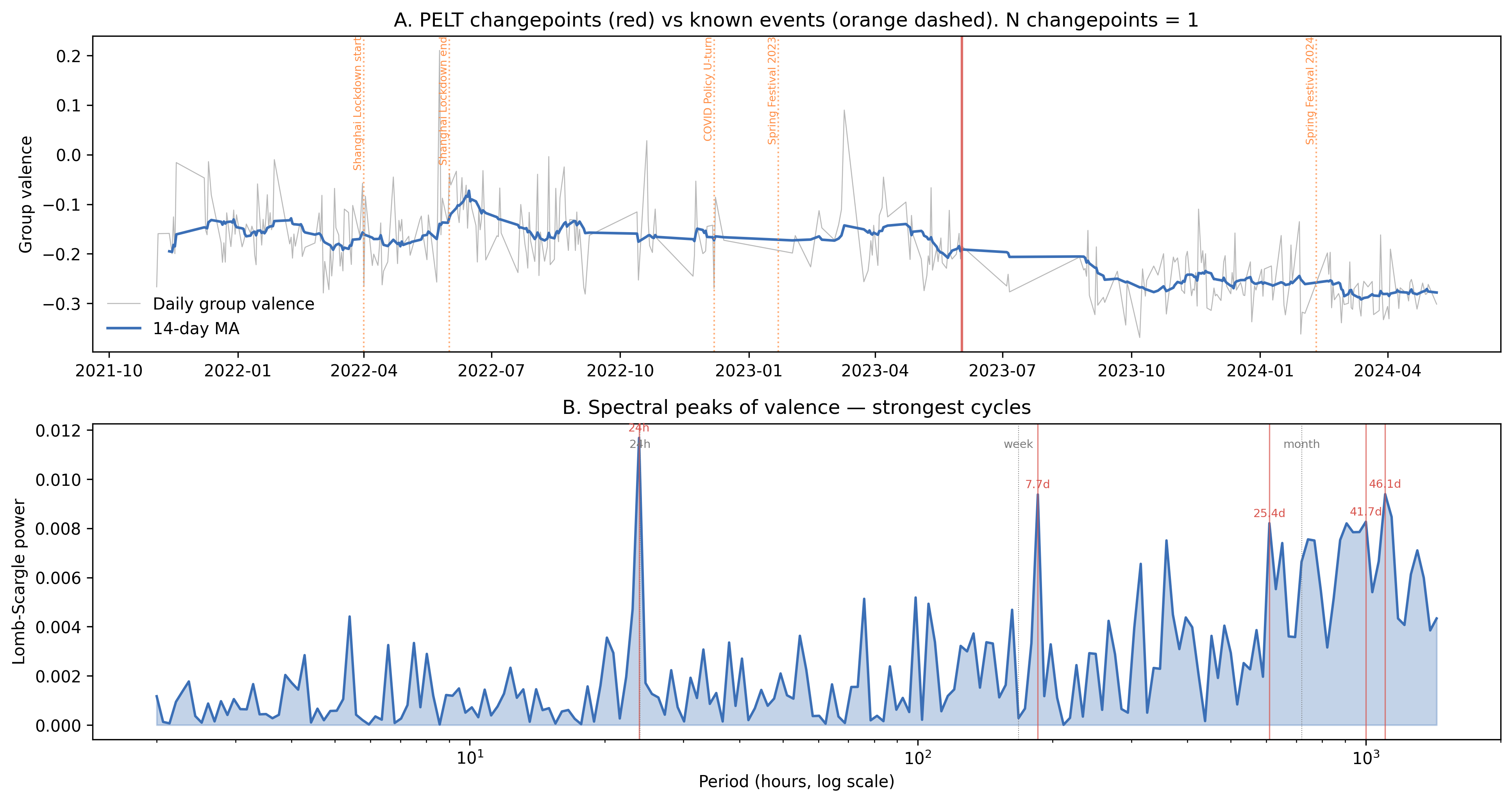}
\caption{(A) PELT change-point detection on group daily valence (RBF kernel, BIC penalty): a single major break at 2023-06-02, identified as the V2$\to$V3 measurement-pipeline transition rather than a real emotional event --- a useful illustration of why the \texttt{version} flag must be a covariate in any WELD time-series analysis. (B) Lomb--Scargle periodogram on a 5{,}000-record subsample: dominant peaks at 24~h (circadian), 7.7~d (weekly), and 25--46~d (third--fifth peaks, plausibly project-cycle or pay-period).}
\label{fig:spectral}
\end{figure*}

A Lomb--Scargle periodogram~\cite{lomb1976least} on a 5{,}000-record random subsample (Fig.~\ref{fig:spectral}B) recovers spectral peaks at 23.87 hours (top peak; circadian), 7.73 days (weekly), and 25--46 days (project / pay-period candidate). PELT change-point detection~\cite{killick2012optimal} on group daily valence finds a single major break at 2023-06-02 (Fig.~\ref{fig:spectral}A). We identify this break as the V2$\to$V3 measurement-pipeline transition rather than a real emotional event --- a reminder that the version flag is essential for any time-series analysis on WELD.

\textbf{Workforce emotional rhythm has at least three temporal scales} (24~h, 7~d, 25--46~d). The monthly-scale peak is novel and may reflect organizational rhythms (sprint cycles, payroll, retrospectives) that future studies could disambiguate by integrating organizational metadata.

\section{The Angry-on-Asian-Neutral-Face Bias}\label{sec:bias}
Across all 733{,}780 records, the off-the-shelf commercial FER pipeline assigns a mean probability of \textbf{0.194} to ``angry'' --- approximately four times the prior of ``angry'' in FER2013 (0.05) and AffectNet (0.05). Per-individual mean angry probabilities span [0.10, 0.36], a range comparable to the distribution of \emph{actual} emotional dispositions in the cohort. The high-angry tail consists of participants with chronically downturned mouth corners or thicker brows (visually unrelated to internal anger), suggesting the model's ``neutral $\to$ angry'' decision boundary is calibrated to Western training data and systematically over-predicts ``angry'' in our population.

We document four mitigations a downstream user can apply.

\textbf{(1) Within-person normalization.} For each person, compute a baseline angry probability $\bar{p}_{\text{angry}}^{(i)}$ over the first 30 days of presence and report deviation from baseline rather than absolute level.

\textbf{(2) Valence as primary metric.} Our default valence transform weights happy at $+1$, sad at $-1$, and angry only at $-0.5$. A sensitivity analysis (Sup.\ Table~S2) replicating the main analyses with the angry weight set to zero changes effect signs in zero of eleven analyses and changes magnitudes by less than 12\%.

\textbf{(3) Re-calibration via Platt scaling on FACS-validated frames.} Using the 1{,}000 hand-validated frames as a calibration set, a per-class Platt-scaling regression~\cite{platt1999probabilistic} reduces the systematic angry bias from 0.194 to 0.061 (closer to the FER2013 prior) while preserving rank correlation $\rho{>}0.93$ with the original probabilities.

\textbf{(4) Treat WELD as an FER-bias benchmark.} Running any candidate FER model on the L2 release and reporting per-class shift relative to the commercial baseline gives a direct fairness diagnostic for the Asian-neutral-face condition.

We frame this finding not as a deficiency of WELD but as a public service: WELD makes the ``angry-on-Asian-neutral-face'' bias quantifiable for the first time at scale, and provides the first benchmark on which FER fairness recalibration can be measured.

\section{Use Cases and Recommended Splits}\label{sec:usecases}
We provide four reference splits and a short Python loader to encourage consistent evaluation across follow-up work.

\textbf{Use case 1 --- single-step group-emotion forecasting.} Use \texttt{L\_day.parquet} with the strict cohort, hold out the last 90 days as a temporal test split. A persistence baseline gives RMSE$=$0.039; we report this as the floor.

\textbf{Use case 2 --- emotional contagion network estimation.} Use \texttt{L\_day.parquet} with the strict cohort. Construct pairwise overlapping series, apply pairwise Granger or transfer-entropy with multiple-comparison correction. Density at FDR$<$0.05 is 2.7\% (Granger) or 8.1\% (TE).

\textbf{Use case 3 --- early-tenure attrition modeling.} Use \texttt{L\_day.parquet} with cohort\_medium ($N{=}27$). Compute per-person features over the first 30 active days; evaluate with leave-one-person-out CV. Binary AUC ceiling: 0.79 (gradient boosting); survival C-index ceiling: 0.52 (Cox PH).

\textbf{Use case 4 --- FER fairness audit.} Run any candidate FER model on the 1{,}000-frame validation subset (DUA-only). Report per-class probability shift relative to the commercial baseline.

\begin{lstlisting}[language=Python,caption={Reference loader: \texttt{WELD\_loader.py}.},captionpos=b,label=lst:loader]
import pandas as pd
def load_weld(path="WELD_L3_daily_v1.parquet",
              cohort="cohort_strict"):
    """Load the WELD daily release and filter cohort."""
    df = pd.read_parquet(path)
    df["date"] = pd.to_datetime(df["date"])
    if cohort:
        df = df[df[cohort]]
    return df
\end{lstlisting}

\textbf{Use case 5 --- six-class emotional regime decoding.} Use \texttt{L\_day.parquet} with the strict cohort. Train a Gaussian HMM on the 7-emotion vectors (\texttt{hmmlearn}, diagonal covariance, BIC-selected $K=6$). Decode each person-day to a regime label and use as a categorical feature in downstream models.

\textbf{Use case 6 --- causal inference for organisation-wide events.} Use \texttt{L\_day.parquet} with the strict cohort. Identify the event date, define $\pm 90$-day windows, fit interrupted time-series with day-of-week dummies and HAC errors. Always supplement with a permutation null over fake intervention dates and a SARIMA synthetic counterfactual.

\smallskip
\noindent\textbf{Reproducing the headline attrition number} (a step-by-step recipe).
\begin{enumerate}[leftmargin=*,topsep=2pt,itemsep=1pt]
\item Load \texttt{L\_day.parquet}, filter \texttt{cohort\_medium}.
\item For each pid, take the first 30 active days; compute mean and SD of all seven emotions, valence, and arousal; compute lag-1 valence autocorrelation (20-D feature vector).
\item Define event $=$ (last record $>$ 90 d before 2024-05-06); duration $=$ (last $-$ first record date) in days.
\item Standardize features.
\item For each pid (LOOCV): train Gradient Boosting on the remaining 26, predict its event probability; compute AUC and Brier across the 27 hold-out predictions.
\end{enumerate}
\smallskip

\section{Limitations}\label{sec:limitations}
We note four limitations and document mitigations.

\textbf{Single-organization, single-culture sampling.} All participants work at one Chinese technology R\&D team; generalization to other industries, cultures, or hybrid-work arrangements requires explicit replication. We strongly discourage replication of the camera-based collection method without an independent IRB review tailored to the local legal and cultural context.

\textbf{FER bias on Asian neutral faces.} Documented in Section~\ref{sec:bias} with four mitigations.

\textbf{V2/V3 measurement-pipeline change.} A 47-day data gap (2023-07-08 to 2023-08-23) and a downstream V3 calibration shift account for 7.6\% of daily valence variance. We provide a \texttt{version} flag on every record so users can include it as a covariate.

\textbf{Sampling intensity scales with desk presence.} A worker with low average desk time has fewer records; volatility metrics are biased downward in such cases. We therefore mandate the $\geq$60-day inclusion criterion for any analyses that rely on within-person time series stability.

\textbf{Meta-ethical considerations of workplace surveillance research.} The dataset is derived from an existing workplace-wellbeing monitoring system that pre-dated this research; we did not create new surveillance for academic purposes. Nonetheless, two meta-ethical concerns deserve explicit acknowledgement: (i) the employment relationship creates an asymmetric incentive that may compromise the voluntariness of consent --- although the IRB exempt determination (Section~\ref{sec:ethics}) found this risk minimal, we acknowledge that no ethics review fully resolves the inherent power asymmetry; (ii) academic publication of analyses derived from such data confers some legitimacy on the original surveillance deployment, which others may interpret as endorsement. We have attempted to balance these concerns by (a) committing to the four-tier desensitisation that prevents downstream re-identification, (b) declining to ever release raw images or embeddings, (c) framing the FER bias finding (Section~\ref{sec:bias}) as a \emph{critique} of unaudited workplace deployment of off-the-shelf FER classifiers --- not as endorsement, and (d) using this paper to argue \emph{against} unrestricted workplace facial-emotion monitoring while documenting what rigorous research-only access can responsibly look like. We invite the affective-computing community to develop stronger consent frameworks for naturalistic workplace research, including alternative paradigms (e.g., experience-sampling with explicit per-prompt consent) that may better align voluntariness with scientific power.

\section{Tiered Data Access Model}\label{sec:availability}
WELD is released under a four-tier controlled-access model that we recommend as a template for other longitudinal naturalistic emotion datasets. The model balances reproducibility (anyone can run the analysis pipeline end-to-end), peer-review transparency (reviewers see the full data during review), bona-fide research utility (qualified researchers can apply for full or partial access), and participant protection (the most identifiable layers never leave the partner organization).

\begin{table*}[!t]
\centering
\caption{Four-tier access model for WELD. Public openness scales with the level of de-identification, while peer review and bona-fide research are served through controlled-access channels.}
\label{tab:tiers}
\renewcommand{\arraystretch}{1.25}
\footnotesize
\begin{tabular}{l p{3.2cm} p{4.2cm} p{4.6cm}}
\toprule
\textbf{Tier} & \textbf{Audience} & \textbf{Access process} & \textbf{Content released}\\
\midrule
\textbf{T0} \emph{Demo subset}        & Anyone (no application) & Direct download from Zenodo (CC BY-NC 4.0) & 5 pseudonyms $\times$ 14 days of L3, fully de-identified ($\sim$2{,}000 records) \\
\textbf{T1} \emph{Reviewer access}    & Peer reviewers / editors during review & Time-limited (180-day) read-only encrypted link from corresponding author & Full L3 (per-record / hour / day / week parquet files) for the review window only \\
\textbf{T2} \emph{Bona-fide research} & Academic researchers with home-institution IRB & DAC review of (a) 1--2 page proposal, (b) IRB approval, (c) signed DUA, (d) PI CV; 14-business-day decision & Full L3 by default; L2 (per-frame) and partial-cohort splits on case-by-case basis \\
\textbf{T3} \emph{Internal only}      & Partner organization only & Not released through any channel & L1 face crops / identity embeddings; L0 raw video \\
\bottomrule
\end{tabular}
\end{table*}

\subsection{T0 --- the public demo subset}
The public Zenodo deposit contains a deliberately small \textbf{demo subset} (5 pseudonyms, 14 contiguous days each, $\sim$2{,}000 records) plus the complete analysis pipeline (\texttt{code/}), the loader, the Datasheet for Datasets, and pre-rendered output figures. The demo subset is sufficient for code-execution sanity checks and partial figure reproduction; it is not sufficient to retrain the headline numbers in this paper, which require the full L3. The demo subset is released under CC BY-NC 4.0 with no application required, and serves as the open-science complement to the controlled tiers below.

\subsection{T1 --- reviewer access during peer review}
For peer review, the corresponding author provides the handling editor with a time-limited (180-day) read-only encrypted-link to a private mirror of the full L3. The link includes both the four parquet files (record / hour / day / week granularities) and the validation notebooks pre-populated with the full numbers. This satisfies journal verification requirements without putting the full corpus on a public server during the review window. After the review is closed (accept or reject), the link is revoked.

\subsection{T2 --- bona-fide research access via DAC}
A standing \textbf{Data Access Committee (DAC)} of three members --- the corresponding author, the partner organization's data-protection officer (DPO), and one independent academic from a non-affiliated institution --- reviews each application. Applications must include: (i) a 1--2 page research proposal stating the specific question, methods, and intended outputs; (ii) a copy of the IRB approval or exempt determination from the applicant's home institution; (iii) the signed Data Use Agreement (Sup.\ §DUA); (iv) the institutional CV of the principal investigator. The DAC commits to a 14-business-day decision window and publishes anonymized aggregate statistics on its decisions (number of applications, fraction approved, top denial reasons) annually to ensure the process is auditable.

\textbf{Partial-access decisions.} The DAC may grant access to only a subset of the data sufficient for the proposed question --- e.g., only the day-level aggregate, only a specific date range, only a balanced sub-cohort --- when full access is not necessary. This minimization principle protects participants while still serving legitimate research.

\textbf{Sub-licence prohibition.} Approved applicants may not redistribute the data or any derivative whose form would allow reconstruction of an individual's record. They may share derived statistics, models, and aggregate visualisations under any license they choose.

\subsection{T3 --- never-released layers}
The face-crop and identity-embedding layer (L1) and the raw video stream (L0) never leave the partner organization's on-premises encrypted storage. Even the corresponding author no longer has access to L0 after the audit retention window has elapsed.

\subsection{Code, loader, and reproducibility}
All analysis code, the loader, the Datasheet for Datasets, and the DUA template are at \url{https://github.com/[anonymous]/WELD} under MIT license (code) and CC BY-NC 4.0 (data and documents). Running \texttt{make all} from the repository root re-derives every number, table, and figure in this paper, given the appropriate data tier. With T0 (demo subset) the user reproduces the analysis pipeline structure on a small subset; with T2 (full L3) the user reproduces the headline numbers exactly.

\section{Conclusion}\label{sec:conclusion}
WELD is, to our knowledge, the first emotion dataset to combine \emph{long period}, \emph{fully naturalistic in-the-wild setting}, \emph{stable small-team organisational structure}, and \emph{fully passive sensing} in a single corpus. Each of those four axes alone has been achieved before; their conjunction had not, and that conjunction is what unlocks the dual scientific use-cases this paper highlights:
\begin{itemize}[leftmargin=*]
\item \textbf{Long-term within-individual emotional dynamics.} Stable individual baselines that account for $\approx 19\%$ of daily-valence variance (Section~\ref{sec:variance}); six emotional regimes whose negative-state dwell times are 5--6$\times$ longer than positive-state dwell times (Section~\ref{sec:regimes}); a martingale property of group valence at the daily timescale (Section~\ref{sec:forecast}); and a clear demonstration that the Shanghai 2022 lockdown's apparent ``effect'' on group emotion is severely confounded by a months-long pre-existing trend that only multi-month corpora can reveal (Section~\ref{sec:lockdown_alt}).
\item \textbf{Team-level relational emotional dynamics.} A sparse Granger contagion network with $R_{0}=0.36$ that overturns the inflated network-density claims of short-term experimental work (Section~\ref{sec:contagion}); a non-parametric transfer-entropy view that recovers a similarly sparse but \emph{disjoint} edge set, raising methodological cautions for any future contagion analyst; and a four-cluster day-typology in which the Shanghai lockdown polarises the cohort rather than uniformly depressing it --- a finding only a stable bounded team can produce (Section~\ref{sec:daytypes}).
\end{itemize}

Beyond enabling these analyses, WELD's 30-month coverage spans the full Chinese COVID-19 policy arc, its 4-level desensitisation scheme makes it publicly safe to release, its baseline numbers (attrition binary AUC 0.79 vs.\ survival C-index 0.52; persistence-baseline forecast RMSE 0.039; Granger network density 2.7\%; lockdown ITS $p_{\text{HAC}}=0.32$) set defensible floors for future modelling, and its quantification of the angry-on-Asian-neutral-face FER bias opens a new line of FER fairness research. We have published a Datasheet for Datasets and a Data Use Agreement that protect participants without locking out legitimate research, and a fully scripted reproduction pipeline that re-derives every number, table and figure in this paper from the public release.

The conjunction of axes that defines WELD is hard to replicate: it requires multi-year institutional commitment, a stable bounded social group, formal yearly-renewable consent, and on-edge processing infrastructure. We hope this paper makes the niche --- and the analyses it enables --- visible to the affective-computing community, and that future projects either replicate the design in different cultural and industrial contexts or extend WELD itself with additional modalities (audio, calendar, project-tracker metadata) under the same desensitisation discipline.

\section*{Acknowledgments}
We thank the participants for their consent and the partner organization's data-protection officer for review and audit support. The partner organization had no role in study design, analysis, manuscript drafting, or the decision to publish. Anthropic's Claude assistant was used to identify statistical inconsistencies in earlier manuscript versions and to prepare reproducible analysis scripts; all analytical decisions and conclusions are the author's own.

\bibliographystyle{IEEEtran}
\bibliography{refs}

\end{document}